\def\year{2022}\relax
\documentclass[letterpaper]{article} 
\usepackage{aaai22}  
\usepackage{times}  
\usepackage{helvet}  
\usepackage{courier}  
\usepackage[hyphens]{url}  
\usepackage{graphicx} 
\usepackage{natbib}  
\usepackage{caption} 
\DeclareCaptionStyle{ruled}{labelfont=normalfont,labelsep=colon,strut=off} 
\usepackage{algorithm}
\usepackage{algorithmic}

\usepackage{newfloat}
\usepackage{listings}

\usepackage{epsfig}
\usepackage{amsmath}
\usepackage{amssymb}
\usepackage{multirow}
\usepackage{bm}
\usepackage{comment}
\usepackage{mathtools}
\usepackage{subcaption}
\usepackage{booktabs}
\usepackage{lipsum}

\newcommand{\ml}[1]{\textcolor{blue}{\textbf{[ML: #1]}}}
\renewcommand{\ml}[1]{}
\newcommand{\cwnote}[1]{\textcolor{cyan}{\textbf{[CW: #1]}}}
\renewcommand{\cwnote}[1]{\crashingcommand}
\newcommand{\cwnotenew}[1]{\textcolor{magenta}{\textbf{[CW: #1]}}}
\renewcommand{\cwnotenew}[1]{\crashingcommand}
\newcommand{\slnote}[1]{\textcolor{red}{\textbf{[SL: #1]}}}
\renewcommand{\slnote}[1]{\crashingcommand}

\renewcommand{\paragraph}[1]{\noindent \textbf{#1}}

\newcommand*\intd{\mathop{}\!\mathrm{d}}

\newcommand{\elbo}{\textsc{ELBO}}
\newcommand{\pointnet}{\textsc{PointNet}}
\newcommand{\vae}{\textsc{VAE}}
\newcommand{\vaes}{\textsc{VAEs}}

\newcommand{\gans}{\textsc{GANs}}
\newcommand{\rgan}{r\textsc{-GAN}}

\newcommand{\treegan}{\textsc{TreeGAN}}
\newcommand{\betavae}{$\beta$\textsc{-VAE}}
\newcommand{\mrgan}{\textsc{MRGAN}}

\newcommand{\pointflow}{\textsc{PointFlow}}
\newcommand{\shapenet}{\textsc{ShapeNet}}

\newcommand{\pointCloud}{\bm X}

\newcommand{\editvae}{\textsc{EditVAE}}

\newcommand{\figref}[1]{Figure~\ref{#1}}
\newcommand{\tabref}[1]{Table~\ref{#1}}

\floatstyle{ruled}
\newfloat{listing}{tb}{lst}{}
\floatname{listing}{Listing}
\title{EditVAE: Unsupervised Parts-Aware Controllable 3D Point Cloud Shape Generation}

\author{
    Shidi Li\textsuperscript{\rm 1},
    Miaomiao Liu\textsuperscript{\rm 1},
    Christian Walder\textsuperscript{\rm 1, \rm 2}
}
\affiliations{
    \textsuperscript{\rm 1}Australian National University\\
    \textsuperscript{\rm 2}Data61, CSIRO\\

    \{shidi.li, miaomiao.liu\}@anu.edu.au, christian.walder@data61.csiro.au
}

\usepackage{bibentry}

\begin{document}

\maketitle

\begin{abstract}
    This paper tackles the problem of parts-aware point cloud generation. Unlike existing works which require the point cloud to be segmented into parts a priori, our parts-aware editing and generation are performed in an unsupervised manner. We achieve this with a simple modification of the Variational Auto-Encoder which yields a joint model of the point cloud itself along with a schematic representation of it as a combination of shape primitives. In particular, we introduce a latent representation of the point cloud which can be decomposed into a disentangled representation for each part of the shape. These parts are in turn disentangled into both a shape primitive and a point cloud representation, along with a standardising transformation to a canonical coordinate system. The dependencies between our standardising transformations preserve the spatial dependencies between the parts in a manner that allows meaningful parts-aware point cloud generation and shape editing. In addition to the flexibility afforded by our disentangled representation, the inductive bias introduced by our joint modeling approach yields state-of-the-art experimental results on the ShapeNet dataset.
\end{abstract}

\section{Introduction}
\label{sect:introdution}
The generation of 3D shapes has broad applications in computer graphics such as automatic model generation for artists and designers~\cite{nash2017shape}, computer-aided design~\cite{mo2020pt2pc} and computer vision tasks such as recognition (Choy et al.~\citeyear{bongsoo2015enriching}). There has been a recent boost in efforts to learn generative shape models from data~\cite{achlioptas2018learning, shu20193d}, with the main trend being to learn the distribution of 3D point clouds using deep generative models such as Variational Auto-Encoders (\vaes )~\cite{cai2020learning, yang2019pointflow}, Generative Adversarial Networks (\gans )~\cite{shu20193d, hui2020progressive, li2021sp}, and normalising flows \cite{yang2019pointflow}. 

Recently,~\citet{mo2020pt2pc} addressed 
structure-aware 3D shape generation, which conditions on the segmentation of point clouds into meaningful \textit{parts} such as the legs of a chair. This yields high quality generation results, but requires time-consuming annotation of the point cloud as a \textit{part-tree} representation. 
A natural alternative therefore, involves extracting a semantically meaningful parts representations in an \textit{unsupervised} manner, using ideas from recent work on \textit{disentangled} latent representations~\cite{chen2018isolating,kim2018disentangling}---that is, representations for which statistical dependencies between latents are discouraged. While disentanglement of the latents allows independent part sampling, reducing the dependence among parts themselves leads to samples with mis-matched style across parts.

\begin{figure}[t]
	\begin{center}
		\includegraphics[width=1\linewidth]{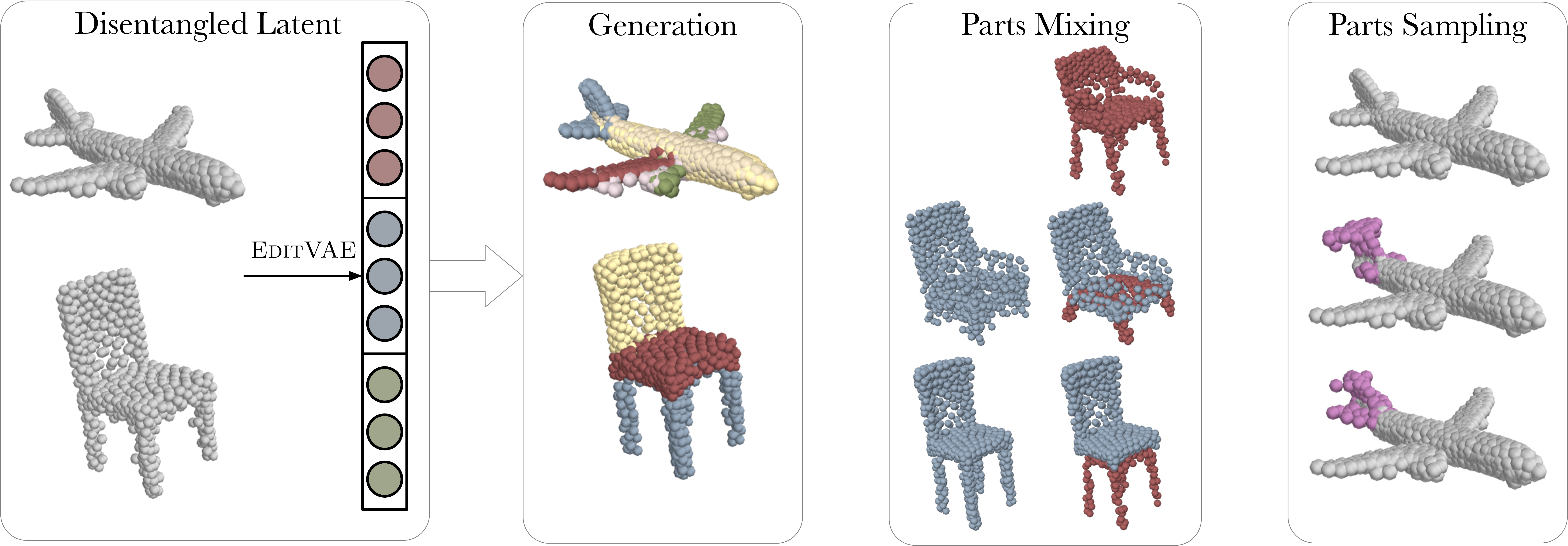}
	\end{center}
	   \caption{Our model learns a disentangled latent representation from point clouds in an unsupervised manner, allowing parts-aware generation, controllable parts mixing and parts sampling. Here we demonstrate: parts-aware \textit{generation} as denoted by the different colours; controllable \textit{parts mixing} to combine the legs of the upper chair with the fixed back and base of the chairs at left; and \textit{parts sampling} of the plane stabilizers.}
	\label{fig:teaser}
\end{figure}

In this paper we propose \editvae, a framework for unsupervised parts-aware
generation. \editvae\ is unsupervised yet learned end-to-end, and allows parts-aware editing while respecting inter-part dependencies. We leverage a simple insight into the \vae\ which admits a latent space that disentangles the style and pose of the parts of the generated point clouds.
Our model builds upon recent advances in primitive-based point cloud representations, to disentangle the latent space into parts, which are modeled by both latent point clouds and latent superquadric primitives, along with latent transformations to a canonical co-ordinate system. 
While we model point-clouds (thereby capturing detailed geometry), our model inherits from the \textit{shape primitive} based point cloud segmentation method of~\citet{paschalidou2019superquadrics}: a semantically consistent segmentation across datasets that does not require supervision in the form of part labeling. 
Given the disentangled parts representation, we can perform shape editing in the space of point-clouds, e.g by exchanging the corresponding parts across point clouds or by re-sampling only some parts.

Our main contributions are summarised as follows. 
\begin{enumerate}
    \item 
    We propose a framework for unsupervised parts-based point cloud generation.
    \item 
    We achieve reliable disentanglement of 
    the latents by modeling points, primitives, and pose for each part.
    \item 
    We demonstrate controllable parts editing via disentangled point cloud latents for different parts.
\end{enumerate}
We provide extensive experimental results on \shapenet\ which quantitatively demonstrates the superior performance of our method as a generator of point clouds.

\section{Related work}\label{sect:related_work}

\paragraph{Disentangled Latent Representation in VAE.}
To promote disentanglement beyond that of the vanilla \vae~\cite{kingma2013auto}, Higgins~et al.~\shortcite{higgins2016beta} introduced an additional KL divergence penalty above that of the usual \textit{evidence lower bound} (\elbo ).
Learning of disentangled latent representations is further investigated by~Kim et. al~\shortcite{kim2018disentangling}, and~\citet{chen2018isolating}.
To handle minibatches while accounting for the correlation of latents,~Kim et. al~\shortcite{kim2018disentangling} proposed a neural-discriminator based estimation while~\citet{chen2018isolating} introduced a minibatch-weighted approximation.
Further, \citet{kim2019relevance} split latent factors into relevant and nuisance factors, treated each in a different manner within a hierarchical Bayesian model~\shortcite{kim2019bayes}.~\citet{locatello2019fairness} showed that disentanglement may encourage fairness with unobserved
variables, and proved the impossibility of learning disentangled representations without inductive biases~\shortcite{locatello2019challenging} in an unsupervised manner, while showing that mild supervision may be sufficient~\shortcite{locatello2019disentangling}.

To learn a reliable disentangled latent representation, the present work introduces a useful inductive bias~\cite{locatello2019challenging} by jointly modeling points, primitives and pose for 3D shapes.
Inspired by the relevance and nuisance factor separation~\cite{kim2019bayes, kim2019relevance}, this work observes and balances the conflict between disentanglement of representation and quality of generation, by separately modeling global correlations of the relative pose of the different parts of a shape, disentangled from their style.
Finally, we fill the gap of learning disentangled latent representations of 3D point cloud in an unsupervised manner, thereby contrasting with much recent disentangled representation learning works focusing on 2D or supervised cases~\cite{kalatzis2020variational, nielsen2020survae,sohn2015learning}.

\paragraph{Neural 3D Point Cloud Generation.}
While 2D image generation has been widely investigated using \gans~\cite{isola2017image,zhu2017unpaired} and \vaes~\cite{kingma2013auto,higgins2016beta,kim2019bayes,sohn2015learning}, neural 3D point cloud abstraction~\cite{tulsiani2017learning, paschalidou2019superquadrics, yang2021unsupervised} and generation has only been explored in recent years.
\citet{achlioptas2018learning} first proposed the \rgan\ to generate 3D point clouds, with fully connected layers as the generator.
In order to learn localized features, Valsesia et al.~\shortcite{valsesia2018learning} and Shu et al.~\shortcite{shu20193d} introduced a generator based on Graph Convolutions.
Specifically, Shu et al.~\shortcite{shu20193d} proposed a tree-based structure with ancestors yielding a neighbor term and direct parents yielding a loop term, named the \treegan .
This design links the geometric relationships between generated points and shared ancestors.
In addition, \pointflow~\cite{yang2019pointflow} learns a distribution of points based on a distribution of shapes by combining \vaes\ and Normalizing Flows~\cite{rezende2015variational}, from which a point set with variable number of points may be sampled.
However, all of the above works generate the point cloud as a whole or by a tree structure without disentanglement, thereby limiting their application power in parts editing.
Although the work by~\citet{chen2019bae} focusing on reconstruction could easily be adapted to unsupervised parts-based generation task, it does not infer precise pose information which is crucial in editing.

A few recent works \cite{nash2017shape,mo2019structurenet,mo2020pt2pc,schor2019componet,dubrovina2019composite, yang2021cpcgan} propose (or could be adapted) to generate point clouds given ground-truth point cloud parts segmentation.
However, the requirement of well-aligned parts semantic labels hinders their real world applications. \mrgan~\cite{gal2020mrgan} firstly attempts to address the parts-aware point
cloud generation by discovering parts of convex shape in an unsupervised fashion. While effective, the decomposed parts may lack semantic meaning. Following this line of work, our \editvae\ approaches parts-aware generation without semantic label requirements. In addition, the proposed model learns a disentangled latent representation, so that the style and pose of parts can be edited independently.
  
\begin{figure*}[pt]
    \begin{center}
       \includegraphics[width=0.8\linewidth]{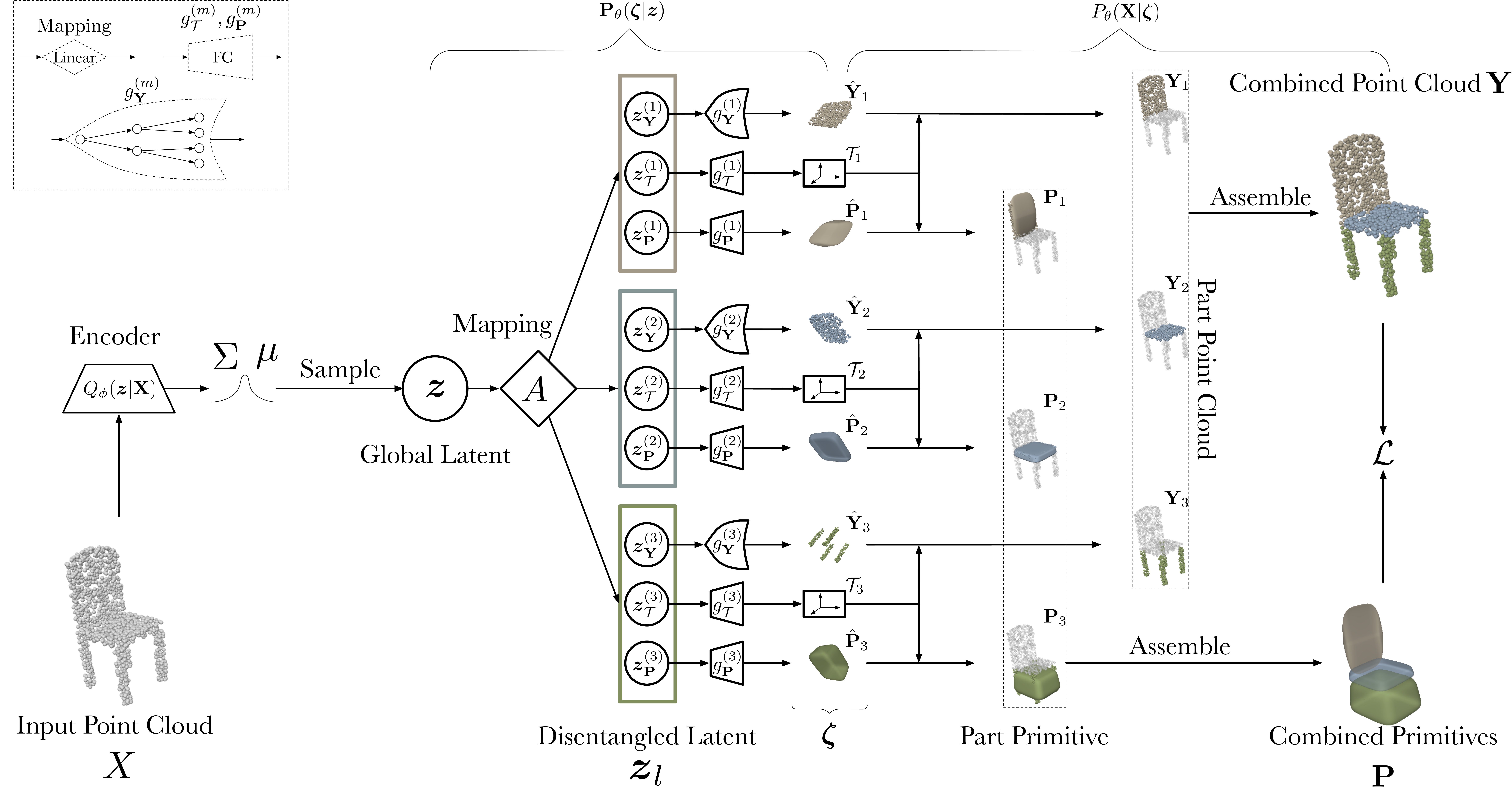}
    \end{center}
       \caption{An overview of the \editvae\ architecture.
       During training, the posterior is inferred by the encoder $Q_\phi$ given the input point cloud $X$, from which a global latent $\bm z$ is sampled.
       The global latent is linearly mapped by $A$ to the disentangled latent $\bm z_l$.
       The disentangled latent maps to parts (denoted by colors), which are further split into point $\hat{\bm Y}_m$, pose $\mathcal T_m$, and primitive $\hat{\bm P}_m$ representations, via the deterministic mappings $g_\star^{(i)}$.
       Each point $\hat{\bm Y}_m$ and primitive $\hat{\bm P}_m$ is transformed to the global coordinate system by the shared pose $\mathcal T_m$.
       The transformed part points $\bm Y_m$ and primitives $\bm P_m$ are then assembled to the complete decoded point cloud $\bm Y$ and primitive $\bm P$ models, respectively.
       Jointly training with a single loss $\mathcal L$ (far right) parsimoniously models key dependencies between point, primitive, and pose models.
       For generation, $\bm z$ is sampled from the standard Gaussian and fed forward to generate point cloud $\bm Y$.
       }
    \label{fig:framework}
\end{figure*}%

\section{Preliminaries}
\label{sect:setups}

To disentangle semantically relevant parts of a 3D point cloud, we decompose it into latent parts which are modeled both as 3D point clouds and 3D shape primitives.

\noindent \paragraph{A point cloud}
in $\mathbb R^{N\times 3}$ is a set of $N$ points sampled from the surface of 3D shape in Euclidean coordinates.

\noindent \paragraph{Primitives} are simple shapes used to assemble parts of more complex shape. 
We employ the superquadric parameterisation for the primitives, which is a flexible model that includes cubes, spheres and ellipsoids as special cases. In line with~\citet{paschalidou2019superquadrics}, we formally  define our superquadric as the two dimensional manifold parameterised by $\eta$ and $\omega$, with surface point
\begin{equation}
    \begin{aligned}
    \bm r(\eta, \omega) = 
    \begin{bmatrix}
    \alpha_x \cos^{\epsilon_1}{\eta} \cos^{\epsilon_2}{\omega} \\
    \alpha_y \cos^{\epsilon_1}{\eta} \sin^{\epsilon_2}{\omega} \\
    \alpha_z \sin^{\epsilon_1}{\eta}
    \end{bmatrix}
    \begin{matrix}
    -\pi/2 \leq \eta \leq \pi/2 \\
    -\pi \leq \omega \leq \pi
    \end{matrix},
    \end{aligned}
\end{equation}
where $\bm \alpha = (\alpha_x, \alpha_y, \alpha_z)^\top$ and $\bm \epsilon = (\epsilon_1, \epsilon_2)^\top$ are the size and shape parameters, respectively. 
 We include additional deformation parameters based on~\citet{barr1987global} in supplementary.

\noindent \paragraph{Pose transformations} are employed to map both the superquadric and point cloud representations of the parts from a canonical pose to the actual pose in which they appear in the \textit{complete} point cloud. We parameterise this transformation as $\bm x \mapsto \mathcal T(\bm x) = \bm{R}(\bm q)\bm x + \bm t$, which is parameterised by a translation  $\bm t \in \mathbb R^3$ and a rotation defined by the quaternion $\bm q \in \mathbb R^4$.
We refer to $\mathcal T$ as the pose for a given part.

\noindent \paragraph{Variational Auto-Encoders ($\vae$)} \cite{kingma2013auto} are an approximate Bayesian inference scheme that introduces an approximate posterior $Q_\phi(\bm z \vert \bm X)$ of the latent representation conditional on the point cloud $\bm X$. The variational parameters $\phi$ are obtained by optimising a bound on the (marginal) data likelihood $P_\theta(\pointCloud)$ known as the \elbo,
\begin{align}
    \log P_\theta(\bm X)
    & \,\,\,\geq\,\,\, \mathbb{E}_{Q_\phi(\bm z \vert \bm X)}[\log P_\theta(\bm X \vert \bm z)] \nonumber \\
    &\quad\,\,\,\,\,\,\,\,-D_{KL}\left( Q_\phi(\bm z \vert \bm X)) \Vert P(\bm z) \right).
\label{eqn:normalelbo}
\end{align}
The first term is known as the reconstruction error, and the second as the variational regulariser. We follow the usual approach of letting the posterior $Q_\phi(\bm z|\bm X)$ be multivariate normal, so that we can employ the usual Monte Carlo approximation with the \textit{reparameterisation trick}~\cite{kingma2013auto} to approximate the reconstruction error. By additionally letting the prior $P(\bm z)$ be multivariate normal, we obtain a closed form expression for the regulariser.

\section{Methodology}\label{sect:methodology}

We motivate our design in next subsection, and then introduce our variational inference scheme, explain how we obtain disentanglement of part latents, give details of the loss functions we use, and conclude with architecture details.

\subsection{Overview of the Design}
\label{sect:methodology:overview}

We divide the challenge of parts-based point cloud generation and editing into the following essential challenges:
\begin{enumerate}
    \item Decomposing multiple (unlabeled) point clouds into semantically meaningful parts.
    \item Disentangling each part into both \textit{style} (such as the shape of the chair leg) and the relative \textit{pose} (the orientation in relation to the other parts of the chair).
    \item Representing the components of the above decomposition in a latent space which allows style and pose to be manipulated independently of one another, while generating concrete and complete point clouds.
\end{enumerate}
We address this problem in an end-to-end manner with a unified probabilistic model. To accomplish this we depart slightly from the well known \vae\ structure, which directly reconstructs the input by the \textit{decoder}. 

For any given input point cloud $\bm X$ we generate a separate point cloud $\hat{\bm Y}_m$ for each part $m$ of the input point cloud (such as the base of a chair), along with a super-quadric prototype $\hat{\bm P}_m$ of that part. This addresses point 1 above. To address point 2, we model $\hat{\bm P}_m$ and $\hat{\bm Y}_m$ in a standardised reference pose via the affine transformation $\mathcal T_m$, and denote by 
\begin{align}
    \bm P_m =\mathcal T_m (\hat{\bm P}_m) \mathrm{~~~~and~~~~}
    \bm Y_m =\mathcal T_m (\hat{\bm Y}_m)
\end{align} 
the point cloud and primitive part representations in the original pose.
 This allows a part's style to be edited while maintaining global coherance. Finally, while we model a single global latent $\bm z$, our decoder generates each part via separate network branches (see \figref{fig:framework}), thereby facilitating various editing operations and satisfying point 3 above.

\subsection{Variational Inference}
\label{sect:methodology:variational}

Our approximate inference scheme is based on that of the \vae~\cite{kingma2013auto,pmlr-v32-rezende14}, but similarly to \citet{kim2019variational} relaxes the assumption that the encoder and decoder map from and to the same (data) space. 
The following analysis is straight-forward, yet noteworthy in that it side-steps the inconvenience of applying variational regularization to $\bm \zeta$.

Denote by $\zeta_m =\{\hat{\bm Y}_m, \hat{\bm P}_m, \mathcal T_m \}$ the $m$-th latent part representation, by $\bm \zeta = \bigcup_{m=1}^M \zeta_m$ the union of all such parts, and by $\bm z$ a global latent which abstractly represents a shape.
We let $Q_\phi(\bm z, \bm \zeta|\bm X)$ represent the approximate posterior with parameters $\phi$, and for simplicity we neglect to notate the dependence of $Q_\phi$ on $\theta$.  Our training objective is the usual marginal likelihood of the data $\bm X$ given the parameters $\theta$,
\begin{align}
    P_\theta( \bm X ) = \int 
    P_\theta( \bm X, \bm z , \bm \zeta) 
    \intd \bm z \intd \bm \zeta.
\end{align}
Taking logs and applying Jensen's inequality we have
\begin{align}
    \hspace{-1.5mm} \log P_\theta(\bm X )
    & = 
    \log \int P_\theta(\bm X, \bm z, \bm \zeta) 
    \intd \bm z \intd \bm \zeta 
    \\
    & = 
    \log \int 
    \frac{Q_\phi(\bm z, \bm \zeta|\bm X)}{Q_\phi(\bm z, \bm \zeta|\bm X)}
    P_\theta(\bm X, \bm z, \bm \zeta ) 
    \intd \bm z \intd \bm \zeta
    \\
    & \geq \int 
    Q_\phi(\bm z, \bm \zeta|\bm X)\log \frac{P_\theta(\bm X, \bm z, \bm \zeta ) }{Q_\phi(\bm z, \bm \zeta|\bm X)}
    \intd \bm z \intd \bm \zeta.
    \label{eqn:hardelbo}
\end{align}
We assume a chain-structured factorisation in our posterior,
\begin{align}
    P_\theta(\bm z, \bm \zeta | \bm X) = P_\theta(\bm \zeta|\bm z) P_\theta(\bm z | \bm X).
\end{align}
Under this factorisation we obtain a tractable variational inference scheme by  assuming that conditional on $\bm z$, the approximate posterior matches the true one, 
\textit{i.e.}
\begin{align}
    Q_\phi(\bm z, \bm \zeta | \bm X)
    & \equiv
    \label{eqn:qexactfactors}
    P_\theta(\bm \zeta | \bm z) \, Q_\phi(\bm z | \bm X).
\end{align}
Putting \eqref{eqn:qexactfactors} into \eqref{eqn:hardelbo} and cancelling $P_\theta(\bm \zeta|\bm z)$ in the $\log$ in \eqref{eqn:hardelbo},
\begin{align}
    \log P_\theta(\bm X)  
    & \,\,\,\geq\,\,\, \mathbb E_{Q_\phi(\bm z |\bm X)}
    \left[
    \log P_\theta(\bm X|\bm \zeta) 
    \right] \nonumber
    \\
    & \quad\,\,\,\,\,\,\,\, -D_{KL}\left( Q_\phi(\bm z | \bm X)) \Vert P_\theta(\bm z) \right),
\end{align}
where $\bm \zeta=\text{NN}_\theta(\bm z)$. \textit{In a nutshell,} this shows that we need only learn an approximate posterior $Q_\phi(\bm z | \bm X)$ via a similar \elbo\ as \eqref{eqn:normalelbo}, to obtain an approximate posterior on $\bm \zeta$. We achieve this via a simple deterministic mapping which, like \citet{nielsen2020survae}, we may notate as the limit $P_\theta(\bm \zeta|\bm z)=Q_\phi(\bm \zeta|\bm z)\rightarrow\delta(\bm \zeta - \mathrm{NN}_\theta(\bm z))$, where $\delta$ is the Dirac distribution and $\mathrm{NN}_\theta$ denotes a neural network. Crucially, while the posterior in $\bm \zeta$ is non-Gaussian, it doesn't appear in the variational regulariser which is therefore tractable.
\subsection{Disentangling the Latent Representation}
\label{sect:methodology:disentanglement}
\editvae\ disentangles the global latent $\bm z$ into a local (to part $\zeta_m$) latent $\bm z_l^{(i)}$, and further to latents for specific component of that part (namely $\bm Y_m, \bm P_m$ or $\mathcal T_m$). We achieve this key feature by \textit{linearly} transforming and partitioning the global latent, \textit{i.e.} we define
\begin{align}
     (\bm z_l^{(1)}, \bm z_l^{(2)}, \cdots, \bm z_l^{(M)})^\top = \bm z_l =  A \bm z,
     \label{eqn:zpartition}
\end{align}
where $A$ is a matrix of weights (representing a linear neural network layer). We further partition the part latents as
\begin{align}
    \bm z_l^{(m)}
    & =
    (\bm z_{\bm Y}^{(m)}, \bm z_{\mathcal T}^{(m)}, \bm z_{\bm P}^{(m)})^\top,
    \label{eqn:locallatents}
\end{align}
and let the corresponding parts themselves be defined as
\begin{align}
    \hat{\bm Y}_m = g_{\bm Y}^{(m)}(\bm z_{\bm Y}^{(m)}),
\end{align}
and similarly for $\hat{\bm P}_m$ and $\mathcal T_m$. Here, $g_{\bm Y}^{(m)}$ non-linearly transforms from the latent space to the part representation.

This achieves several goals. First, we inherit from the \vae\ a meaningful latent structure on $\bm z$. Second, by \textit{linearly} mapping from $\bm z$ to the local part latents $\bm z_{\bm Y}^{(i)}, \bm z_{\mathcal T}^{(i)}$ and $ \bm z_{\bm P}^{(i)}$, we ensure that linear operations (\textit{e.g.} convex combination) on the global latent precisely match linear operations on the local latent space, which therefore captures a meaningfully \textit{local} latent structure. Finally, partitioning $\bm z_l$ yields a representation that disentangles parts by construction, while dependencies between parts are captured by $A$. Experiments show we obtain meaningful disentangled parts latents.
\begin{table*}[t]
    \centering
    \caption{Generative performance.
    $\uparrow$ means the higher the better, $\downarrow$ means the lower the better.
    The score is highlighted in bold if it is the best one compared with state-of-the-art.
    Here $M$ is the number of minimum parts we expect to separate in training.
    For network with $\star$ we use the result reported in~\cite{valsesia2018learning,shu20193d}}
    \begin{tabular}{c c c c c c c} 
     \toprule
     Class & Model & JSD $\downarrow$ & MMD-CD $\downarrow$ & MMD-EMD $\downarrow$ & COV-CD $\uparrow$ & COV-EMD $\uparrow$ \\ 
     \midrule
     \multirow{7}{*}{Chair} & \rgan\ (dense)$^\star$ & 0.238 & 0.0029 & 0.136 & 33 & 13 \\ 
      & \rgan\ (conv)$^\star$ & 0.517 & 0.0030 & 0.223 & 23 & 4 \\
      & Valsesia (no up.)$^\star$ & 0.119 & 0.0033 & 0.104 & 26 & 20 \\
      & Valsesia (up.)$^\star$ & 0.100 & 0.0029 & 0.097 & 30 & 26 \\
      & \treegan\ \cite{shu20193d} & 0.119 & 0.0016 & 0.101 & 58 & 30 \\
      & \mrgan\ \cite{gal2020mrgan} & 0.246 & 0.0021 & 0.166 & \textbf{67} & 23 \\
      & \editvae\ (M=7) & \textbf{0.063} & \textbf{0.0014} & \textbf{0.082} & 46 & \textbf{32} \\
      & \editvae\ (M=3) & \textbf{0.031} & 0.0017 & 0.101 & 45 & \textbf{39} \\
      \midrule
      \multirow{7}{*}{Airplane} & \rgan (dense)$^\star$ & 0.182 & 0.0009 & 0.094 & 31 & 9 \\ 
      & \rgan (conv)$^\star$ & 0.350 & 0.0008 & 0.101 & 26 & 7 \\
      & Valsesia (no up.)$^\star$ & 0.164 & 0.0010 & 0.102 & 24 & 13 \\
      & Valsesia (up.)$^\star$ & 0.083 & 0.0008 & 0.071 & 31 & 14 \\
      & \treegan\ \cite{shu20193d} & 0.097 & \textbf{0.0004} & 0.068 & 61 & 20 \\
      & \mrgan\ \cite{gal2020mrgan} & 0.243 & 0.0006 & 0.114 & \textbf{75} & 21 \\
      & \editvae\ (M=6) & \textbf{0.043} & \textbf{0.0004} & \textbf{0.024} & 39 & \textbf{30} \\
      & \editvae\ (M=3) & \textbf{0.044} & 0.0005 & \textbf{0.067} & 23 & 17 \\
     \midrule
     \multirow{3}{*}{Table} & \treegan\ \cite{shu20193d} & 0.077 & 0.0018 & 0.082 & 71 & \textbf{48} \\
      & \mrgan\ \cite{gal2020mrgan} & 0.287 & 0.0020 & 0.155 & \textbf{78} & 31 \\
      & \editvae\ (M=5) & 0.081 & \textbf{0.0016} & \textbf{0.071}  & 42 & 27 \\
      & \editvae\ (M=3) & \textbf{0.042} & \textbf{0.0017} & 0.130 & 39 & 30 \\
     \bottomrule
    \end{tabular}
    \label{table:generative}
\end{table*}%
\subsection{Loss Functions}
\label{sect:methodology:losses}

Completing the model of the previous sub-section requires to specify the log likelihood $\log P_\theta(\bm X|\bm \zeta)$, which we decompose in the usual way as the negative of a sum of loss functions involving either or both of the point $\bm Y_m$ and super-quadric $\bm P_m$, representations---combined with the standardisation transformation $\mathcal T$ which connects these representations to the global point cloud, $\bm X$. Note that from a Bayesian modelling perspective, there is no need to separate the loss into terms which decouple $\bm P$ and $\bm Y$; indeed, the flexibility to couple these representations within the loss is a source of useful inductive bias in our model. 

While our loss does not correspond to a normalised conditional $P_\theta(\bm X|\bm \zeta)$, working with un-normalised losses is both common \cite{sun2019learning,paschalidou2019superquadrics}, and highly convenient since we may engineer a practically effective loss function by combining various carefully designed losses from previous works.

\paragraph{Point Cloud Parts Loss.}
We include a loss term for each part point cloud $\hat{\bm Y}_m$ based on the Chamfer distance
\begin{align}
    & \mathcal{L}_c(\bm X, \bm Y)
    = \\
    & \frac{1}{2 \vert \bm X \vert } \sum_{x \in \bm X} \min_{y \in \bm Y} \Vert x - y \Vert_2^2
    + 
    \frac{1}{2 \vert \bm Y \vert } \sum_{y \in \bm Y} \min_{x \in \bm X} \Vert x - y \Vert_2^2.
    \nonumber
\end{align}
We sum over parts to obtain a total loss of
\begin{align}
    \mathcal L_{\bm Y} = \sum_{m=1}^M \, \mathcal{L}_c(\hat{\bm X}_m, \hat{\bm Y}_m),
\end{align}
where $\bm X_m$ is the subset of $\bm X$ whose nearest superquadric is $\bm P_m$, and $\hat{\bm X}_m = \mathcal T^{-1}(\bm X_m)$ is in canonical pose.

\paragraph{Superquadric Losses.}
The remaining terms in our loss relate to the part $\bm P_m$ and combined $\bm{P}=\bigcup_{m=1}^M \bm P_m$ primitives, and would  match \citet{paschalidou2019superquadrics} but for the addition of a regulariser which discourages overlapping superquadrics, \textit{i.e.}\footnote{$\mathcal{L}_{o}(\bm{P})$ matches the implementation of~\citet{paschalidou2019superquadrics} provided by the authors.}
\begin{align}
        &\mathcal{L}_{o}(\bm{P}) \\
        &= \frac{1}{M} \smash{\sum_{m=1}^M}
        \frac{1}{\vert \bm{S} \vert - \vert \bm{S_m} \vert} \sum_{\substack{\bm s \in \bm{S} \setminus \bm{S}_m }} \max\left(1-H_m(\bm s), 0\right) 
        ,
        \nonumber
\end{align}
where $\vert \cdot \vert$ denotes cardinality, $\bm S_m$ is a point cloud sampled from $\bm P_m$, $\bm S=\bigcup_{m=1}^M \bm S_m$, and $H_m(\cdot)$ is the smoothed indicator function for $\bm P_m$ defined in~\citet{solina1990recovery}. 
  
\begin{figure*}[t]
    \begin{center}
       \includegraphics[width=0.9\linewidth]{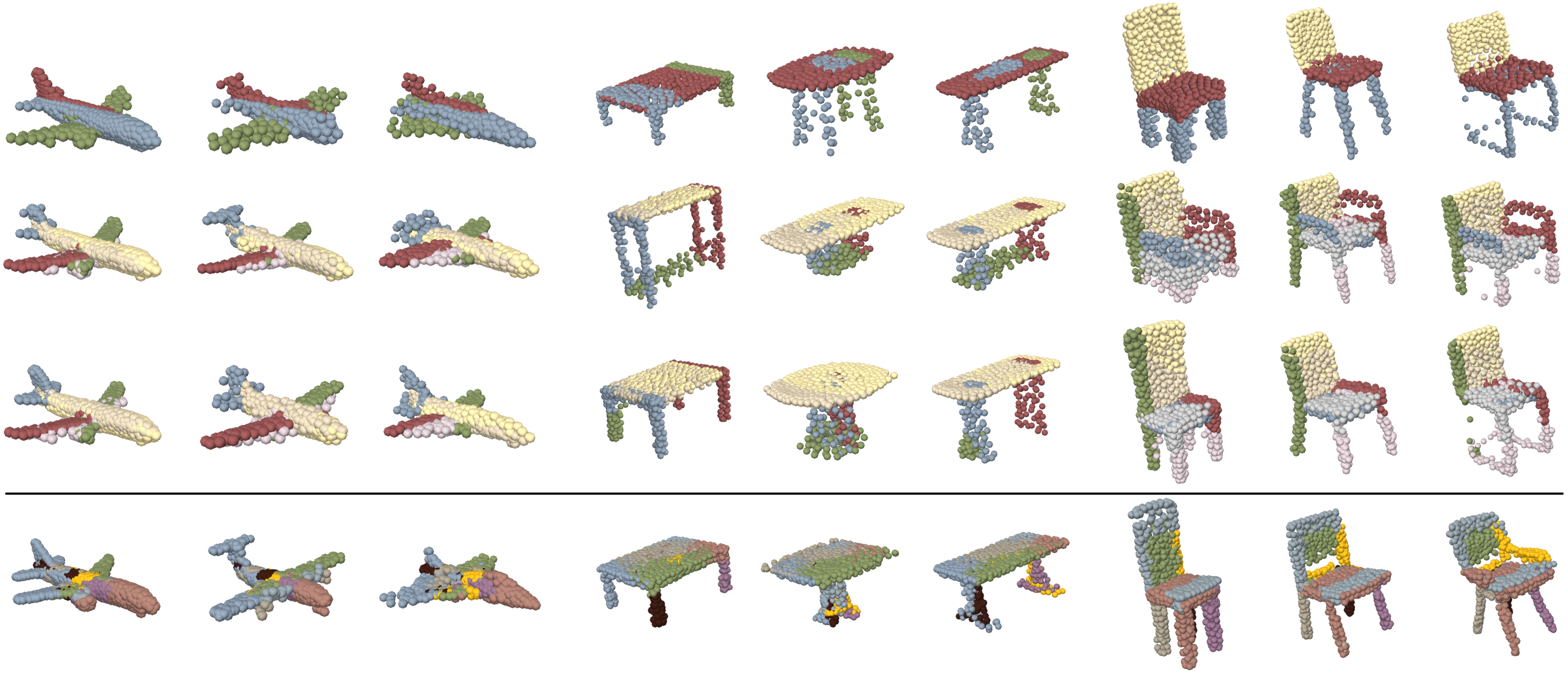}
    \end{center}
       \caption{Parts-based generated point clouds from the airplane, table and chair categories, coloured by part.
       \textit{Bottom row:} examples generated by \treegan~\cite{shu20193d}.
       The top three rows are \editvae---the top row with $M=3$, and the second and third rows with the number of parts $M$ reported in Table~\ref{table:generative}.
       }
    \label{fig:generation}
\end{figure*}%

\subsection{Architecture Details}
\label{sect:methodology:architecture}

\editvae\ framework is shown in \figref{fig:framework}. The posterior $Q_{\phi}({\bm z}|{\bm X})$ is based on the $\pointnet$ architecture \cite{qi2017pointnet}, with the same structure as~\citet{achlioptas2018learning}. 
For $P_\theta(\bm \zeta|\bm z)$, we apply the linear transform and partitioning of \eqref{eqn:zpartition} for disentangled part representations followed by further shape and pose disentanglement. We use the generator of \treegan~\cite{shu20193d} as the decoder, modelling ${g^{(i)}_{\bm Y} }$, to generate the point cloud for each part. The super-quadric decoder modules match \citet{paschalidou2019superquadrics} for primitive generation ${\bm P}_m$, as do those for the $\mathcal T_m$.
Weights are not shared among branches.

\section{Experiments}\label{sect:experiment}
\paragraph{Evaluation metrics.} 
We evaluate our \editvae\ on the ShapeNet~\cite{shapenet2015} with the same data split as~\citet{shu20193d} and report results on the three dominant categories of chair, airplane, and table.  We adopt the evaluation metrics of~\citet{achlioptas2018learning}, including Jensen-Shannon Divergence (JSD), Minimum Matching Distance (MMD), and Coverage (COV). 
As MMD and COV may be computed with either Chamfer Distance (CD) or Earth-Mover Distance (EMD), we obtain five different evaluation metrics, \textit{i.e.} JSD,\, MMD-CD,\, MMD-EMD,\, COV-CD,\, and COV-EMD.

\paragraph{Baselines.} We compare with four existing models of \rgan~\cite{achlioptas2018learning}, Valsesia~\cite{valsesia2018learning}, $\treegan$~\cite{shu20193d} and \mrgan~\cite{gal2020mrgan}. \rgan~and Valsesia generate point clouds as a single whole without parts inference or generation based on a tree structure as in \treegan. Similar to our approach, \mrgan\ performs unsupervised parts-aware generation, but with ``parts'' that lack a familiar semantic meaning and without disentangling pose.

\paragraph{Implementation details.}\footnote{Code will be provided on publication of the paper.}  The input point cloud consists of a set of 2048 points, which matches the above baselines. Our prior on the global latent representation ${\bm z}\in \mathbb{R}^{256}$ is the usual standard Gaussian distribution. 
We chose $\bm z_{\bm Y}^{(m)}\in \mathbb{R}^{32}$, and $\bm z_{\mathcal T}^{(m)}$, $\bm z_{\bm P}^{(m)}\in \mathbb{R}^8$ for the local latents of \eqref{eqn:locallatents}. 
We trained \editvae\ using the \textsc{Adam} optimizer~\cite{kingma2014adam} with a learning rate of $0.0001$ for 1000 epochs and a batch size of 30. 
To fine-tune our model we adopted the \betavae\ framework~\cite{higgins2016beta}.

\subsection{Results}\label{subsect:generation}
\paragraph{Generation.} \editvae\ generates point clouds by simply sampling from a standard Gaussian prior for $\bm z$, mapping by $A$ and and the subsequent part branches of Figure~\ref{fig:framework}, before merging to form the complete point cloud. We show quantitative
and qualitative results in \tabref{table:generative} and \figref{fig:generation}, respectively. As shown in \tabref{table:generative}, the proposed \editvae\  achieves competitive results (see \textit{e.g.} the $M=7$ results for the chair category) compared with the states of the art.
The parts number $M$ is manually selected to achieve a meaningful semantic segmentation, \textit{e.g.} a chair may be roughly decomposed into back, base, and legs for $M=3$.
Furthermore, while~\citet{shu20193d} generates point clouds according to a tree structure---and could thereby potentially generate points with consistent part semantics---it does not allow the semantics-aware shape editing due to lacking of disentangled parts representations. To the best of our knowledge, \mrgan~\cite{gal2020mrgan} is the only other method achieving parts-disentangled shape representation and generation in an unsupervised manner. The results in \tabref{table:generative} show that our method outperforms MRGAN in both the JSD and MMD metrics. Morover, \editvae\ achieves highly semantically meaningful parts generation as shown in \figref{fig:generation} and the experiment as discussed below, which further achieves parts-aware point cloud editing.

\paragraph{Parts Editing.} 
\editvae\ disentangles the point clouds into latents for each part, and then in turn into the point cloud, pose, and primitive for each part.
This design choice allows editing some parts with other parts fixed, yielding controllable parts editing and generation.
We demonstrate this via both parts mixing and parts (re-)sampling. 

\begin{figure}[t]
    \begin{center}
       \includegraphics[width=0.8\linewidth]{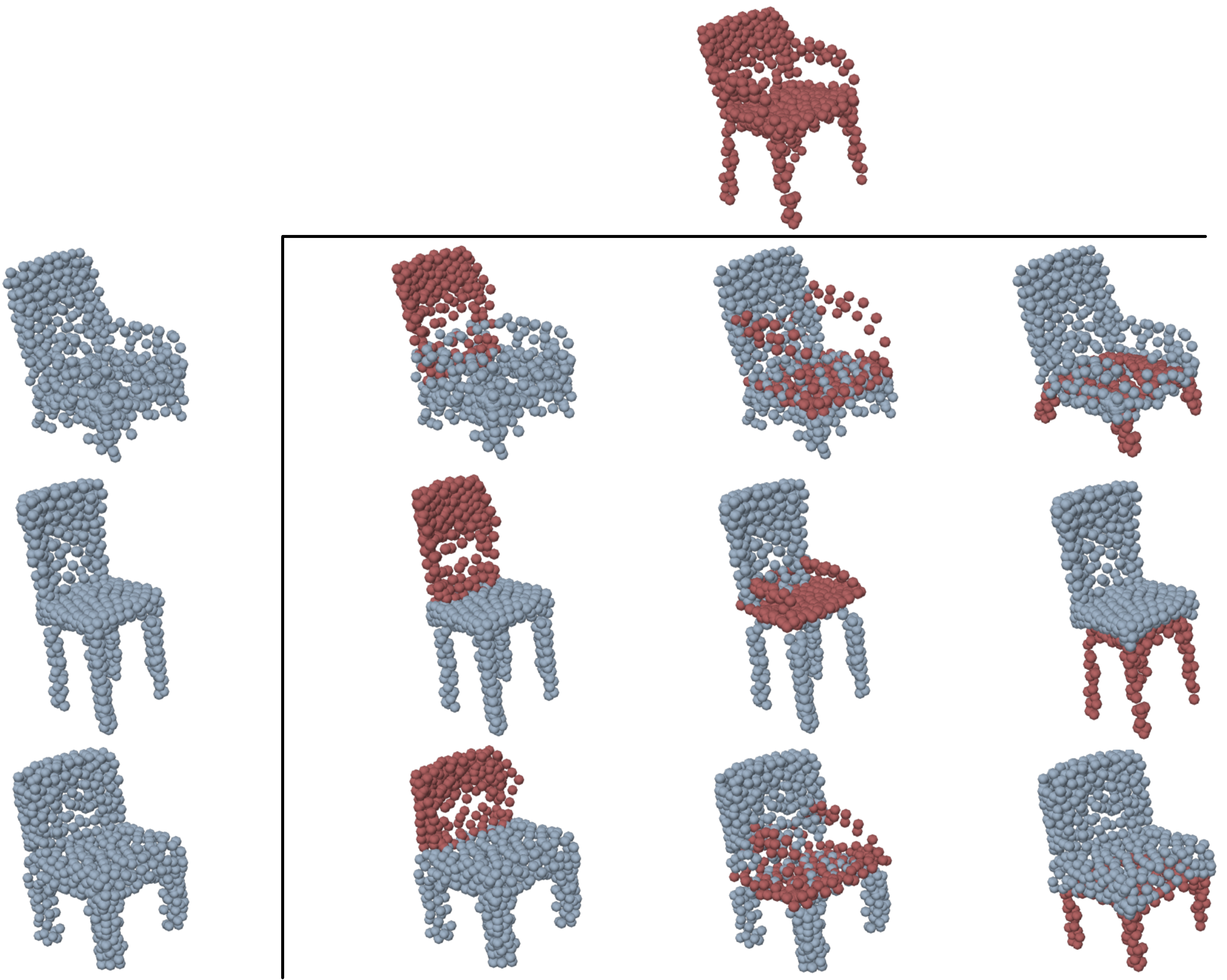}
    \end{center}
       \caption{Parts mixing in the chair category with $M=3$. \textit{Far left:} ground truth point clouds, \textit{top:} reference point cloud. \textit{Remaining:} from left to right, back, base, and legs for ground truth points are mixed by corresponding parts in the reference one via mixing their disentangled latents.
       }
    \label{fig:parts_mixing}
\end{figure}%
\begin{table*}[h]
    \centering
    \caption{Semantic meaningfulness measurements. $M=*$ represents \editvae\ in\tabref{table:generative}. The lower MCD the better.
    }
    \begin{tabular}{c|ccc|ccc|ccc} 
    \toprule
    \multirow{2}{*}{Model} & \multicolumn{3}{c|}{Chair} & \multicolumn{3}{c|}{Airplane} & \multicolumn{3}{c}{Table} \\
     \cline{2-10}
     & \textsc{TreeGAN} & M=3 & M=7 & 
      \textsc{TreeGAN} & M=3 & M=6 &
      \textsc{TreeGAN} & M=3 & M=5\\
    \midrule
    MCD$\downarrow$ & 0.0164 & 0.0028 & 0.0121 & 0.0043 & 0.0016 & 0.0018 & 0.0266 & 0.0121 & 0.0214 \\
    \bottomrule
    \end{tabular}%
    \label{table:measurement}
\end{table*}
\begin{table}[t]
    \centering
    \caption{Generative performance for the entire shape and its parts, for the chair category.
    Semantic labels are obtained by primitive segmentation in our framework.
    }
    \begin{tabular}{c|c| c c c} 
     \toprule
    \multirow{2}{*}{Model} & \multicolumn{4}{c}{MMD-CD$\downarrow$} \\
    \cline{2-5}
    & as whole & base & back & leg\\
    \midrule
     \editvae\  & 0.0017 & 0.0016 & 0.0014 & 0.0024 \\
     \textsc{Baseline}\ & 0.0025 & 0.0017 & 0.0015 & 0.0024\\
    \bottomrule
    \end{tabular}%
    \label{table:generative_editing}
\end{table}%
\begin{table}[t]
    \centering
    \caption{Generative performance comparsion for \editvae\ and two baselines in chair category.
    }
    \resizebox{\columnwidth}{!}{%
    \begin{tabular}{c c c c c c} 
     \toprule
     Model & JSD$\downarrow$ & MMD-CD$\downarrow$ & COV-CD$\uparrow$ \\
    \midrule
    \emph{Baseline-G} & 0.062 & 0.0019 & 42 \\
    \emph{Baseline-S} & 0.163 & 0.0030 & 10 \\
    \editvae\ (M=3) & 0.031 & 0.0017 & 45 \\
    \editvae\ (M=7) & 0.063 & 0.0014 & 46 \\
    \bottomrule
    \end{tabular}%
    }
    \label{table:robust_gen}
\end{table}%
\paragraph{Parts Mixing.}
\label{subsubsect:parts_mixing}
It is defined by exchanging some parts between generated reference and ground-truth point clouds while keeping others fixed.
We achieve mixing by transferring corresponding parts latents from reference to ground-truth, and further transforming it by the generator and pose of the parts in the ground-truth.
The corresponding part in the ground-truth point cloud may therefore be changed to the style of the reference one. For example, the results in the first row of \figref{fig:parts_mixing} show that the ground-truth shape of a sofa with solid armed base may be changed into a larger hollow armed one based on its reference shape with consistent style. Namely, the size and pose of mixed parts follow that of the ground-truth, but keep the style from the reference.

\paragraph{Parts Sampling.}
This involves resampling some parts in a generated point cloud.
For resampled parts, we fix the pose but resample the point cloud parts latent. The
fixed pose is essential to maintain generated part point clouds with a consistent location that matches the other fixed parts to achieve controllable generation.
\begin{figure}[t]
    \begin{center}
       \includegraphics[width=0.8\linewidth]{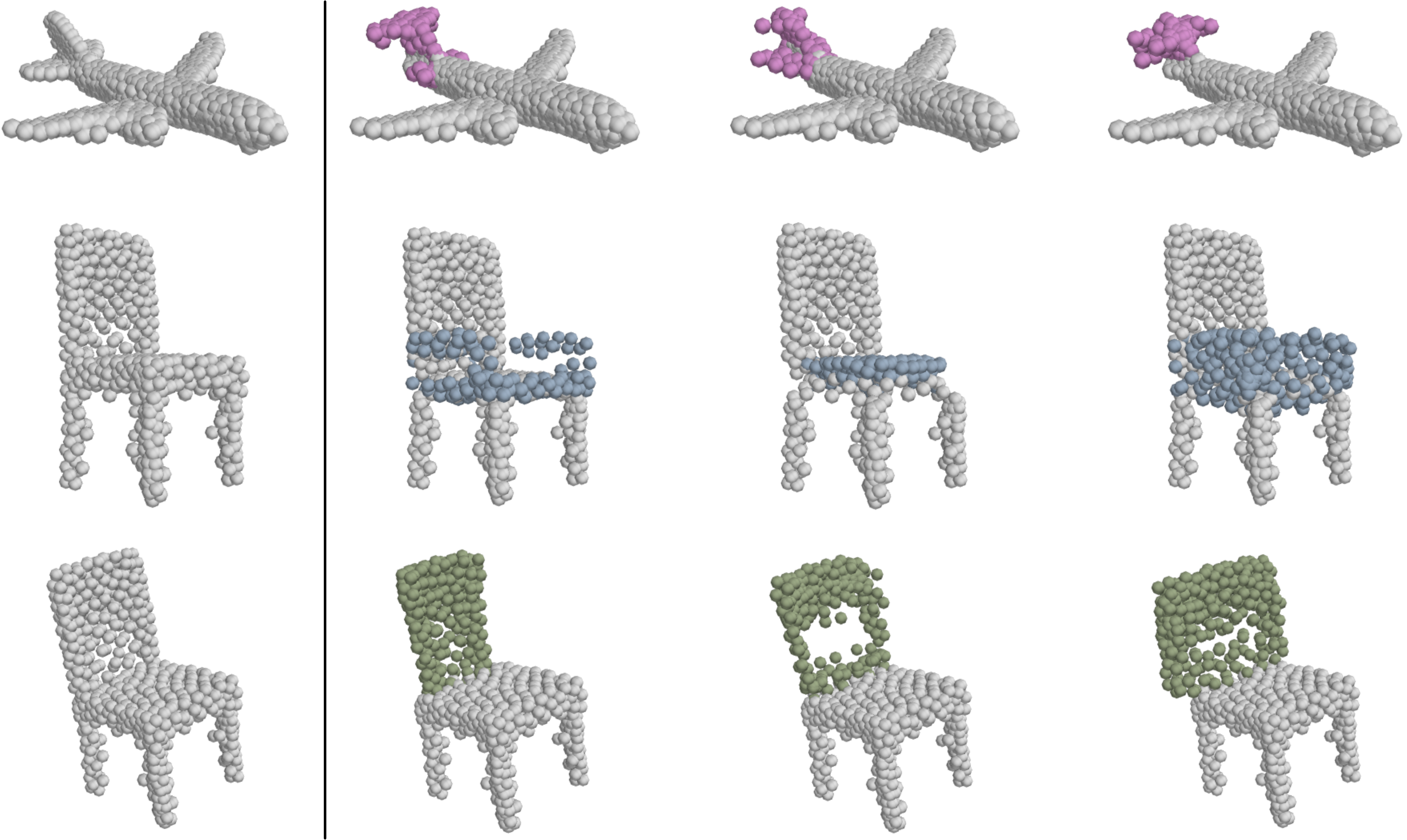}
    \end{center}
       \caption{Parts sampling. \textit{Far left:} the reference point clouds. Colored parts in the three right columns are sampled from latent space---from top to bottom, we sampled the airplane stabilizer, chair base, and chair back.}
    \label{fig:parts_sampling}
\end{figure}%
Qualitative results for parts sampling are in \figref{fig:parts_sampling}.
Variations in the part styles demonstrated the controllable point cloud generation.

\paragraph{Semantic Meaningfulness.}
We first define a vanilla measurement by comparing the distance between the ground truth semantic label and the unsupervisedly generated one.
The distance is defined as the mean of smallest Chamfer distance for each unsupervised part with respect to all ground truth parts (MCD in \tabref{table:measurement}).
As \mrgan~\cite{gal2020mrgan} lacks accompanying code, we mainly compare the semantic meaningfulness with respect to \treegan\ in \tabref{table:measurement}.
\editvae\ outperforms when we define the ground truth segmentation as the most meaningful.

\subsection{Ablation Studies}

\paragraph{Generation / Editing Trade-Off.}
 We aim to evaluate the influence
of the linear mapping $A$ for disentangled
representation learning (see \figref{fig:framework}). To this end, we introduce a \textsc{Baseline} framework by simply removing this $A$. Results are shown in
 \tabref{table:generative_editing}. Specifically, we compare 
our generation with the \textsc{Baseline} results at the whole
point cloud level and at the parts level, such as the \emph{base}, \emph{leg}, and \emph{back}, for the 
chair category. While \textsc{Baseline} achieves disentangled parts-aware representation learning and comparable results for parts sampling to \editvae\footnote{We evaluate each part generation result separately.}, the manner in which \textsc{Baseline} generates points as a whole via sampling from a standard Gaussian yields inferior performance due to the mismatched style across parts.
Thus, the mapping $A$ manages to decouple the undesirable generation / editing trade-off caused by disentanglement.
Detailed analysis and visualizations are in the supplementary materials.

\paragraph{Stage-wise Baselines.}
We compared \editvae\ with two stage-wise baselines defined as 
 \emph{Baseline-S} and \emph{Baseline-G}. 
In particular, \emph{Baseline-S} is built by first generating parts labels via the state-of-the-art unsupervised segmentation method~\cite{paschalidou2019superquadrics} followed by a supervised parts-aware generation  approach~\cite{schor2019componet}.
\emph{Baseline-G} is created
by training the the point cloud branch in \figref{fig:framework} with the ground-truth parts segmentation.
The comparison is performed on the chair category in \shapenet~\cite{shapenet2015}, and reported in \tabref{table:robust_gen}. 

\editvae\ is robust to semantic segmentation as its generation is close to \emph{Baseline-G}.~Further, the performance of $M=3$ is closer to \emph{Baseline-G} compared with $M=7$, in line with our observation (see \figref{fig:generation}) that this case achieves a similar segmentation to the ground-truth.
Further,~\editvae\ outperforms \emph{Baseline-S} by overcoming the style-mismatch issue and is robust to noise introduced by mapping parts to a canonical system with learned poses.

\section{Conclusions}\label{sect:conclustions}
We introduced \editvae , which generates parts-based point clouds in an unsupervised manner. 
The proposed framework learns a disentangled latent representation with a natural inductive bias that we introduce by jointly modeling latent part- and pose-models, thereby making parts controllable.
Through various experiments, we demonstrated that \editvae\ balances parts-based generation and editing in a useful way, while performing strongly on standard pointcloud generation metrics.

\section{Acknowledgement}
Miaomiao Liu was supported in part by the Australia Research Council DECRA Fellowship (DE180100628) and ARC Discovery Grant (DP200102274).

\appendix

\bibliography{aaai22}



\clearpage
\appendix

\section{Generation / Editing Trade-off Analysis \& Results}
\label{sect:ablation}
\begin{table*}[t]
    \centering
    \caption{More results in generation/editing trade-off}
    \begin{tabular}{c|c|c|c| c c c c c c c} 
     \toprule
     \multirow{2}{*}{Class} & \multirow{2}{*}{Primitive Number} & \multirow{2}{*}{Model} & \multicolumn{8}{c}{MMD-CD $\downarrow$}  \\ 
     \cline{4-11}
     & & & as whole & part A & part B & part C & part D & part E & part F & part G \\
     \midrule
     \multirow{4}{*}{Chair} & \multirow{2}{*}{7} & \textsc{EditVAE}\ & \textbf{0.0014} & 0.0012 & 0.0011 & 0.0015 & 0.0013 & 0.0025 & 0.0015 & 0.0013 \\
     & & \textsc{Baseline}\ & 0.0029 & 0.0014 & 0.0012 & 0.0019 & 0.0014 & 0.0027 & 0.0016 & 0.0015 \\
     \cline{3-11}
     & \multirow{2}{*}{3} & \textsc{EditVAE}\ & \textbf{0.0017} & 0.0014 & 0.0016 & 0.0024 & - & - & - & -\\
     & & \textsc{Baseline}\ & 0.0025 & 0.0016 & 0.0016 & 0.0024 & - & - & - & - \\
      \midrule
    
     \multirow{4}{*}{Airplane} & \multirow{2}{*}{6} & \textsc{EditVAE}\ & \textbf{0.0004} & 0.0004 & 0.0005 & 0.00004 & 0.0006 & 0.0006 & 0.0005 & - \\
     & & \textsc{Baseline}\ & 0.0007 & 0.0004 & 0.0005 & 0.0005 & 0.0006 & 0.0007 & 0.0005 & - \\
     \cline{3-11}
     & \multirow{2}{*}{3} & \textsc{EditVAE}\ & \textbf{0.0005} & 0.0006 & 0.0005 & 0.0007 & - & - & - & - \\
     & & \textsc{Baseline}\ & 0.0006 & 0.0006 & 0.0005 & 0.0008 & - & - & - & - \\
     \midrule
    
     \multirow{4}{*}{Table} & \multirow{2}{*}{5} & \textsc{EditVAE}\ & \textbf{0.0016} & 0.0020 & 0.0011 & 0.0023 & 0.0015 & 0.0020 & - & - \\
     & & \textsc{Baseline}\ & 0.0042 & 0.0024 & 0.0011 & 0.0030 & 0.0016 & 0.0022 & - & - \\
     \cline{3-11}
     & \multirow{2}{*}{3} & \textsc{EditVAE}\ & \textbf{0.0017} & 0.0025 & 0.0012 & 0.0022 & - & - & - & -\\
     & & \textsc{Baseline}\ & 0.0035 & 0.0034 & 0.0013 & 0.0025 & - & - & - & - \\
     \bottomrule
    \end{tabular}
    \label{table:getf}
\end{table*}%
We aim to evaluate the influence
of the linear mapping $A$ for disentangled
representation learning (see Figure 2 in the main paper). To this end, we introduce a \textsc{Baseline} framework by simply removing this $A$. 
Results are shown in the main paper Table 3.
Specifically, we compare our generation with the \textsc{Baseline} results at the whole point cloud level and at the parts level, 
such as the \emph{base}, \emph{leg}, and \emph{back}, for the chair category. 
While \textsc{Baseline} achieves disentangled parts-aware representation learning and comparable results for parts sampling to \editvae\, 
the manner in which \textsc{Baseline} generates points as a whole via sampling from a standard Gaussian yields inferior performance due to the mismatched style across parts.

We observe that well-disentangled latents benefit controllable editing, as we may unilaterally alter the style of one part, 
without affecting that of the other parts. 
This is mainly due to our particular disentangled representation which discourages certain dependencies among latents. 
By contrast, parts-based generation requires strong correlation within latent factors to generate style-matched point clouds. 
Hence, this disentanglement is fundamentally opposing to the parts-based point cloud generation as a whole due to the lack of global correlation across parts.

This observation can be further explained by the concept of relevant and nuisance latents separation in~\cite{kim2019relevance} which addresses the balance between reconstruction and
generation. Specifically, relevant latents depend on the input and vice versa, which indicates that the global ``style'' information is stored in the relevant latent. 
Completely disentangled latents can achieve perfect reconstruction, as the known inputs can lead to fully observed relevant and nuisance latents. 
However, relevant latents are randomly sampled in generation due to the lack of input as observation. 
As a result, disentangled latents with different ''style'' information lead to a style mismatch across the generated part point clouds. 
We thus introduce a linear mapping $A$ to encode the ''relevant'' latents consistently across disentangled part latents, to achieve parts-aware generation with a consistent style.

We provide more quantitative results in Table~\ref{table:getf}.
Similar to the results reported in Table 1 of the main paper, we compare the generative performance of \textsc{EditVAE}\ with a \textsc{Baseline} for which we removed the linear mapping $A$ from our model.
\begin{figure}[t]
    \begin{center}
       \includegraphics[width=1\linewidth]{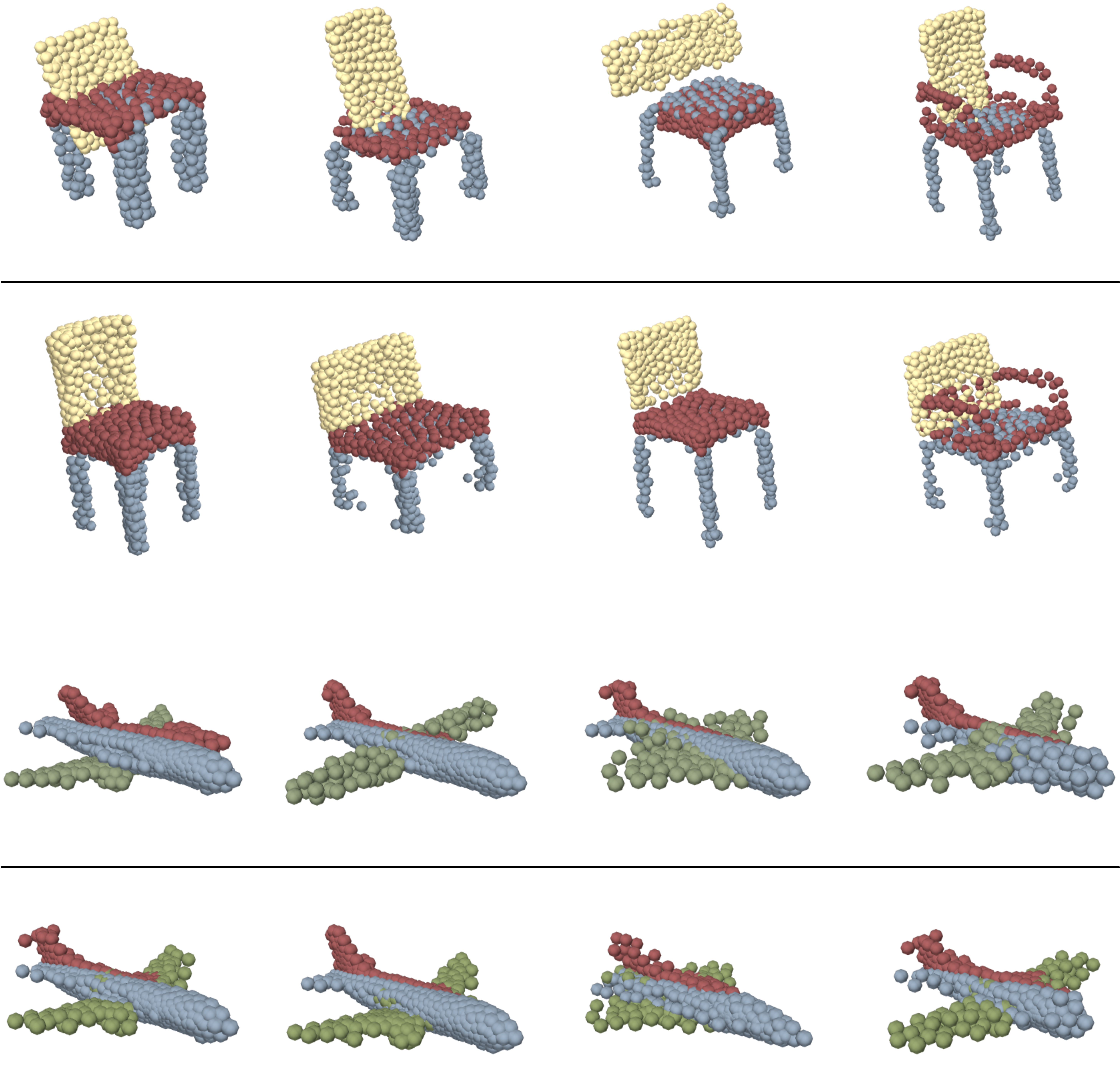}
    \end{center}
       \caption{Visualization of point clouds generated by \textsc{EditVAE}\ (below lines) 
       and \textsc{Baseline}\ (above lines).
       Colors denote the different parts.}
    \label{fig:trade_off}
\end{figure}%
As shown in Table~\ref{table:getf}, the proposed \textsc{EditVAE}\ consistently outperforms the \textsc{Baseline}\, for all three categories and for various numbers of primitives $M$.
The quantitative results demonstrate that sampling from disentangled latents without global context information leads to point clouds of low quality. 
More qualitative comparison results are provided in Figure~\ref{fig:trade_off}, which shows that the style and pose are mismatched in general among parts for point clouds generated by \textsc{Baseline}.
For example, back parts in the chair category either intersect  the base (left most), or are detached from it (third column).
In addition, the back sizes are also not matched to the bases (all four examples).
For airplanes generated by \textsc{Baseline}, we observe glider's wings (middle left) and fighter's wings (middle right) being assembled with civil airliners.
Moreover, as sampled pose latents are mismatched with sampled point latents, the stabilizers are added at the wrong position (left most). 

In summary, the `style' of parts is mismatched in point clouds generated by \textsc{Baseline}, mainly because the disentangled latents do not keep the global correlations within parts.
By contrast, our model can generate point clouds in a consistent style due to our global context-aware latents disentanglement which is achieved by the linear mapping $A$ in our framework.

\section{Additional Mixing Examples} \label{sect:mixing} 
In the main paper we showed parts mixing results for the chair category in \textsc{ShapeNet}~\cite{shapenet2015} with number of primitives $M=3$. Here we will provide more parts mixing results on other categories.

\begin{figure}[t]
    \begin{center}
       \includegraphics[width=1\linewidth]{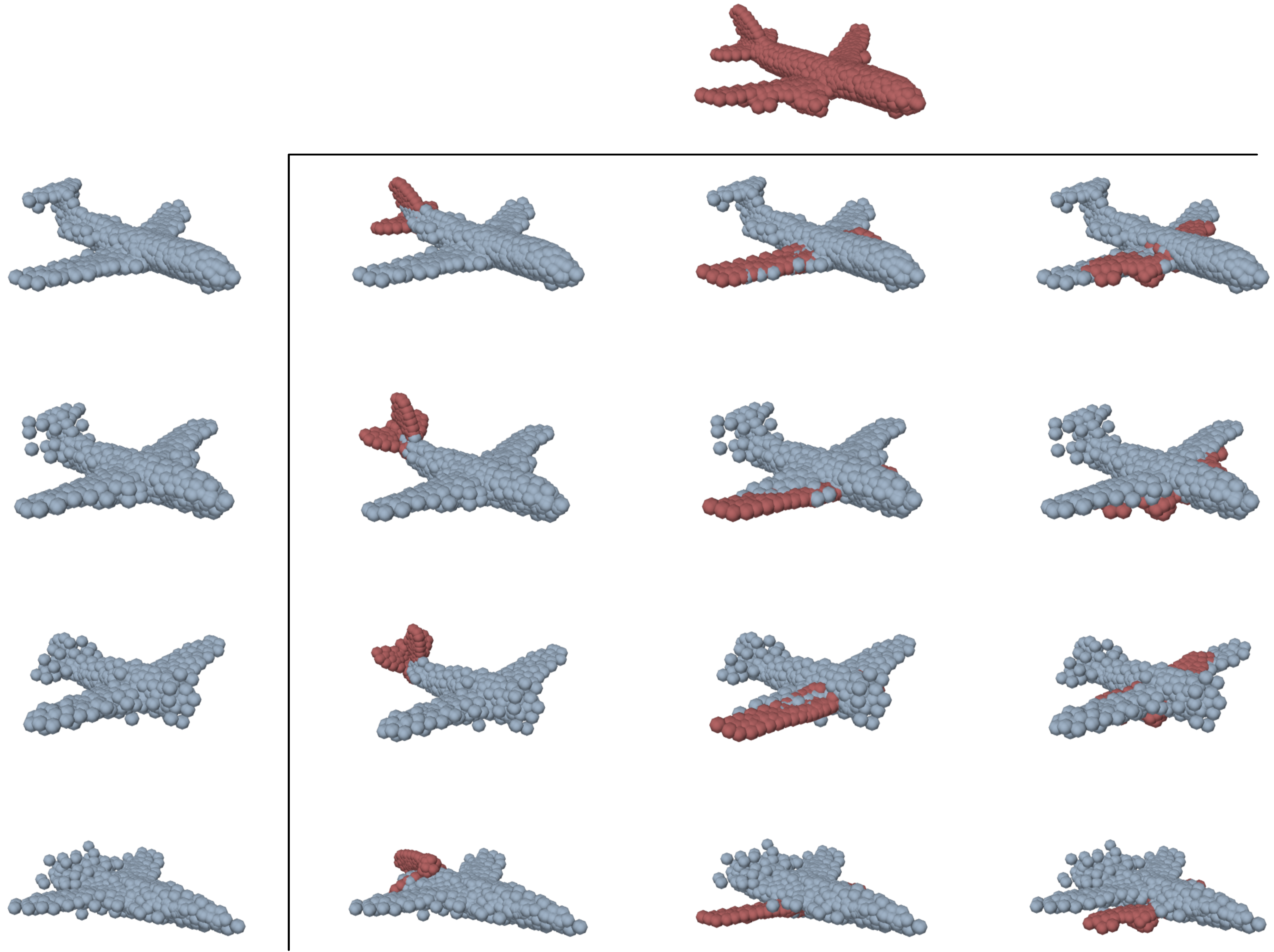}
    \end{center}
       \caption{Parts mixing in the airplane category with $M=6$. 
       \textit{Far left:} ground truth point clouds, \textit{top:} reference point cloud.
       \textit{Remaining:} from left to right: stabilizer, right wing, and engine of the ground truth point clouds are replaced by corresponding ones in the reference via mixing of their disentangled latents.
       }
    \label{fig:parts_mixing_airplane7}
\end{figure}
\begin{figure}[t]
    \begin{center}
       \includegraphics[width=1\linewidth]{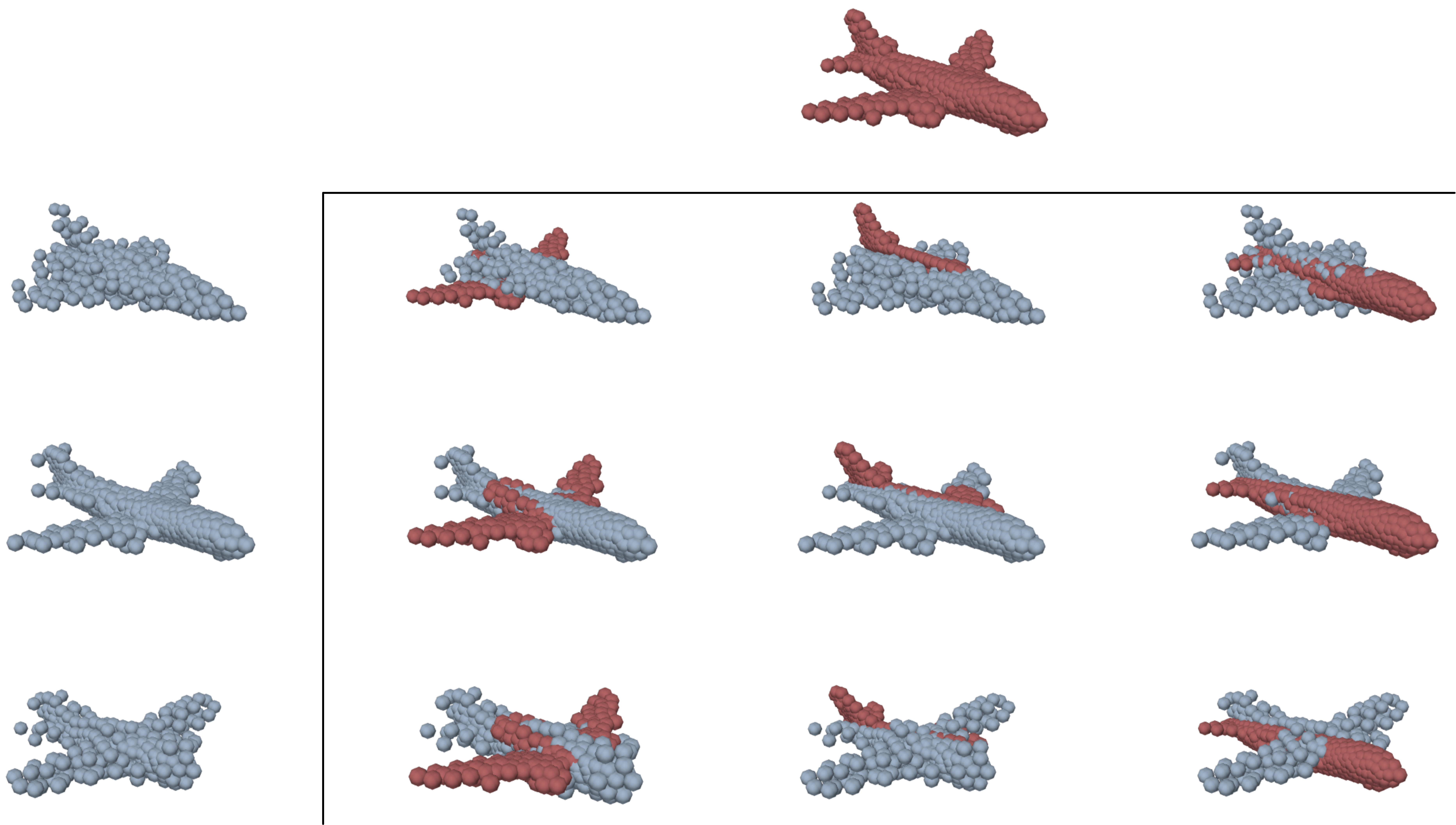}
    \end{center}
       \caption{Parts mixing in the airplane category with $M=3$. 
       \textit{Far left:} ground truth point clouds, \textit{top:} the reference point cloud.
       \textit{Remaining:} from left to right: the wings, stabilizer, and body for ground truth points are replaced by the corresponding parts in the reference one via mixing their disentangled latents.
       }
    \label{fig:parts_mixing_airplane3}
\end{figure}

%
In Figure~\ref{fig:parts_mixing_airplane7}, we mix parts in the airplane category with number of primitives $M=6$.
Each ground truth point cloud (blue) is mixed with a reference  point cloud (red) with respect to the stabilizer, the right wing, and the engine.
In the first column of Figure~\ref{fig:parts_mixing_airplane7}, the shapes of all stabilizers in the ground truth point clouds are changed to that of the reference one but respecting their poses, which leads to a mixed point cloud with consistent style.
In addition, the ground truth airplanes without engines are also `assembled' with reference's engine by the mixing operation.
It is worth noting that the style of remaining parts has not been changed thanks to our learned disentangled representation.
Similar observations can be found in Figure~\ref{fig:parts_mixing_airplane3}.
\begin{figure}[t]
    \begin{center}
       \includegraphics[width=1\linewidth]{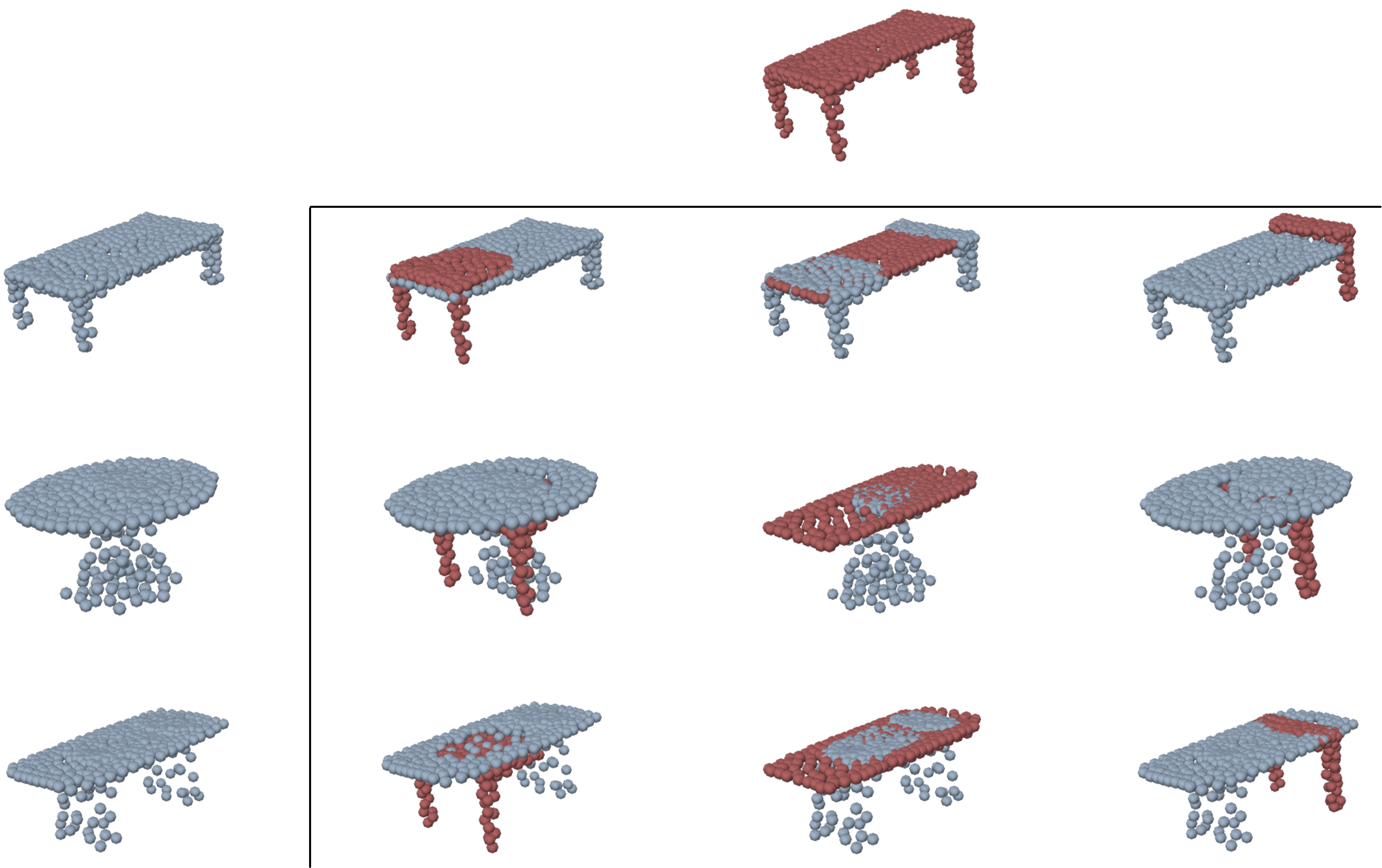}
    \end{center}
       \caption{Parts mixing in the table category with $M=3$. 
       \textit{Far left:} ground truth point clouds, \textit{top:} reference point cloud.
       \textit{Remaining:} from left to right: right legs, left legs, and base for ground truth points are replaced by the corresponding parts in the reference one via mixing of the disentangled latents.
       }
    \label{fig:parts_mixing_table}
\end{figure}%
We additionally show our mixing results on the table category 
in Figure~\ref{fig:parts_mixing_table}. As demonstrated in the figure, we can change the round base of the table to a rectangular one from the reference point cloud in a consistent style.

\section{Additional Sampling Examples}\label{sect:sampling}
\begin{figure}[t]
    \begin{center}
       \includegraphics[width=1\linewidth]{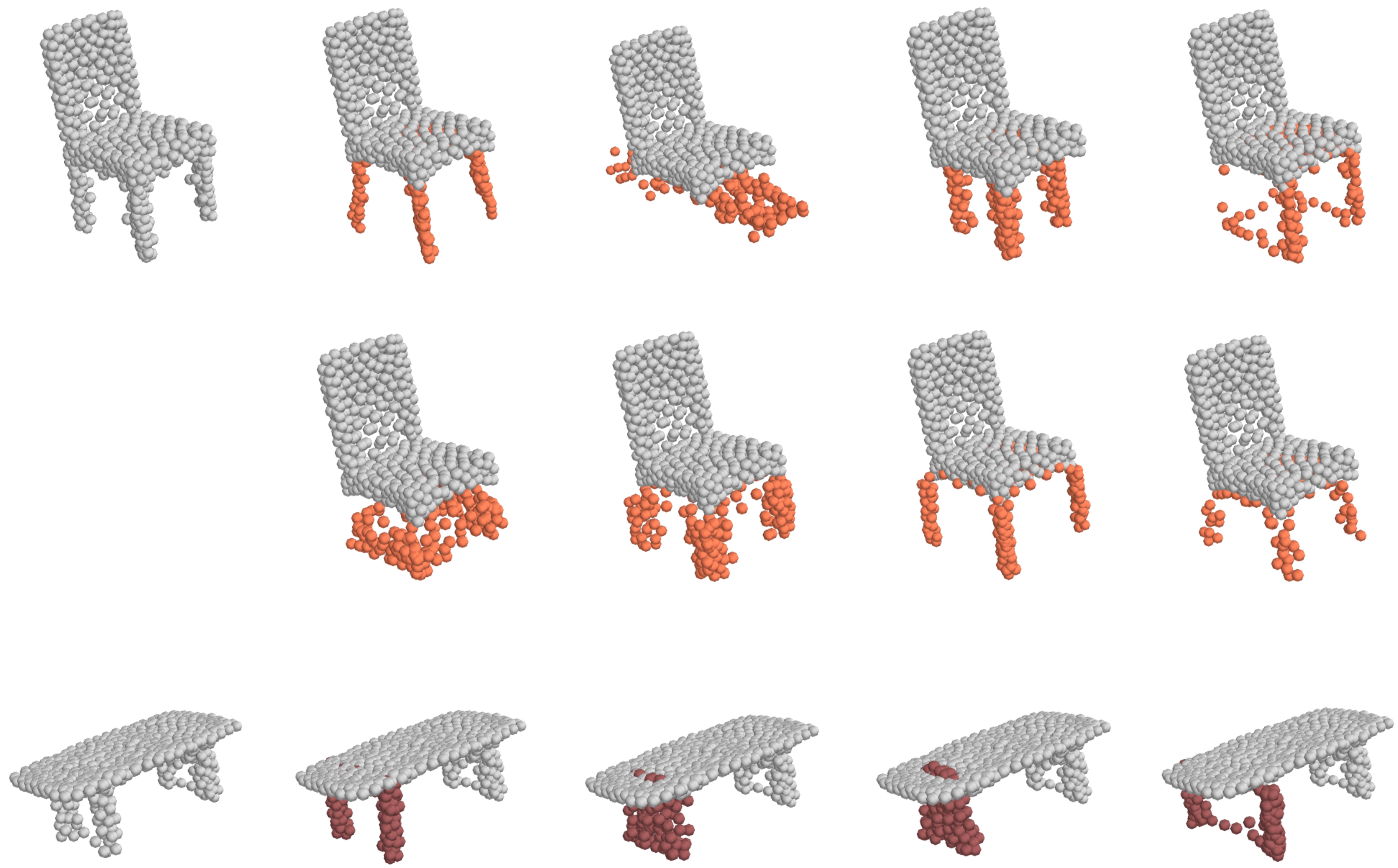}
    \end{center}
       \caption{Parts sampling. 
       \textit{Far left:} the reference point clouds. 
       Colored parts in the three right columns are sampled from the latent space
       ---from top to bottom, we sampled the chair legs and table legs.}
    \label{fig:parts_sampling}
\end{figure}%
As the parts distribution is unknown, we achieve parts sampling by first sampling a global latent from a multivariate normal distribution and then passing to the linear mapping $A$.
Another option is passing the parts latent to a Real NVP layer~\cite{dinh2016density} before feeding to the generators/decoders during training.
By letting Real NVP learn to map the parts latent into a standard normal distribution, we may then generate novel parts by sampling the parts latent directly.
Both options are equivalent if the Real NVP layer is linear, as it can be included in generators/decoders. 
In order to have a simple and elegant model, we removed the Real NVP layer in the main paper.

\begin{figure}[t]
    \begin{center}
       \includegraphics[width=1\linewidth]{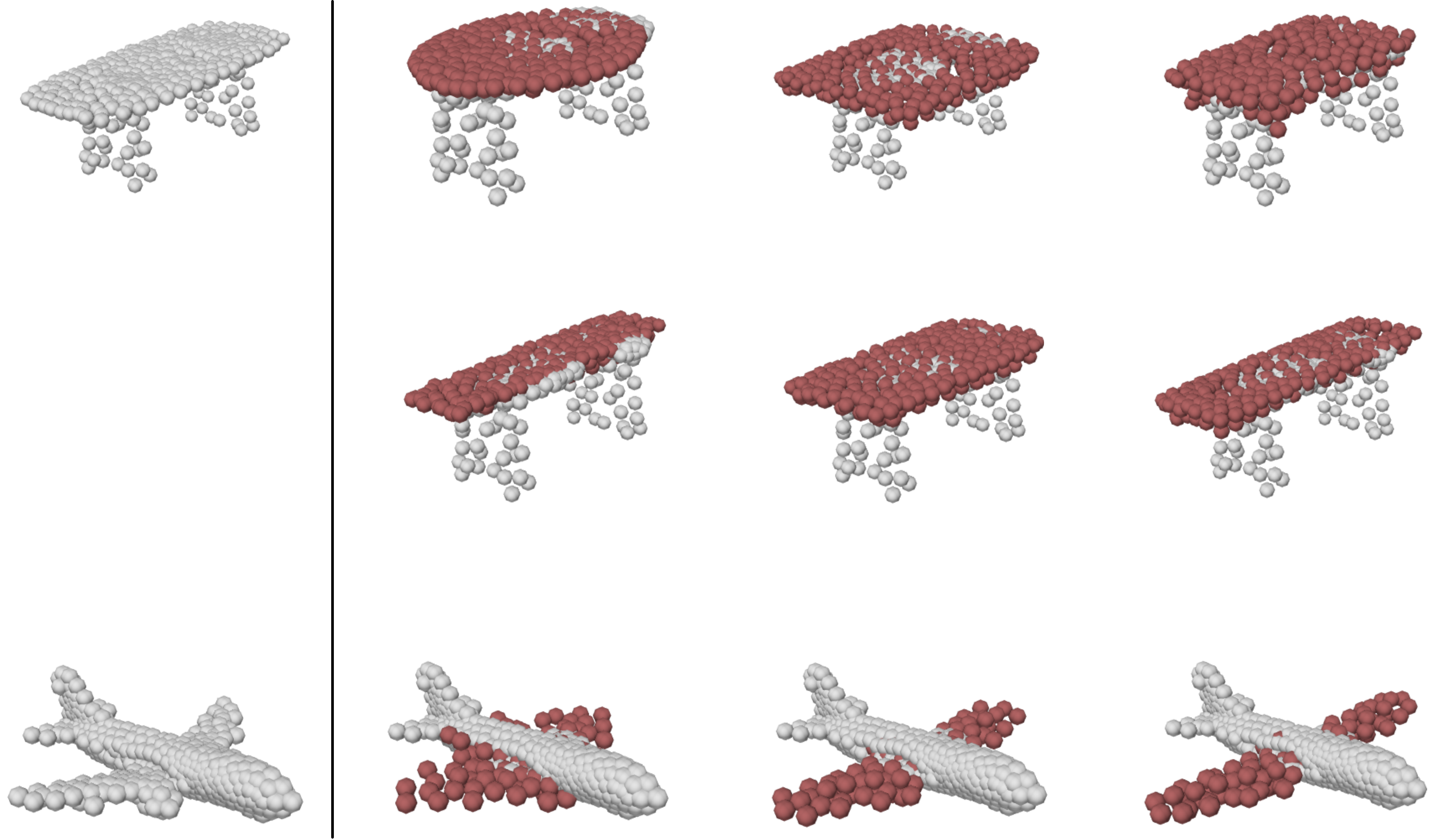}
    \end{center}
       \caption{Parts sampling. 
       \textit{Far left:} the reference point clouds. 
       Colored parts in the three right columns are sampled from the latent space
       ---from top to bottom, we sampled the table base and airplane wings.}
    \label{fig:parts_sampling2}
\end{figure}

Additional parts sampling results may be found in Figures~\ref{fig:parts_sampling} and~\ref{fig:parts_sampling2}.
We sampled chair legs and table right legs in Figure~\ref{fig:parts_sampling}.
In particular, different styles (normal or sofa style), sizes (thick or slim), and pose (rotation) of legs are sampled from our disentangled latents. 
Moreover, we provide more results for parts sampling of table bases and airplane wings in Figure~\ref{fig:parts_sampling2}.

As shown in the figure, different shapes of table base (round, rectangular and square), and styles of airplane wing (glider's and fighter's wing) are sampled while the remaining parts are held fixed. We see that parts sampling allows us to achieve controllable point clouds generation.

\section{Interpolation}
\begin{figure}[h]
    \begin{center}
       \includegraphics[width=1\linewidth]{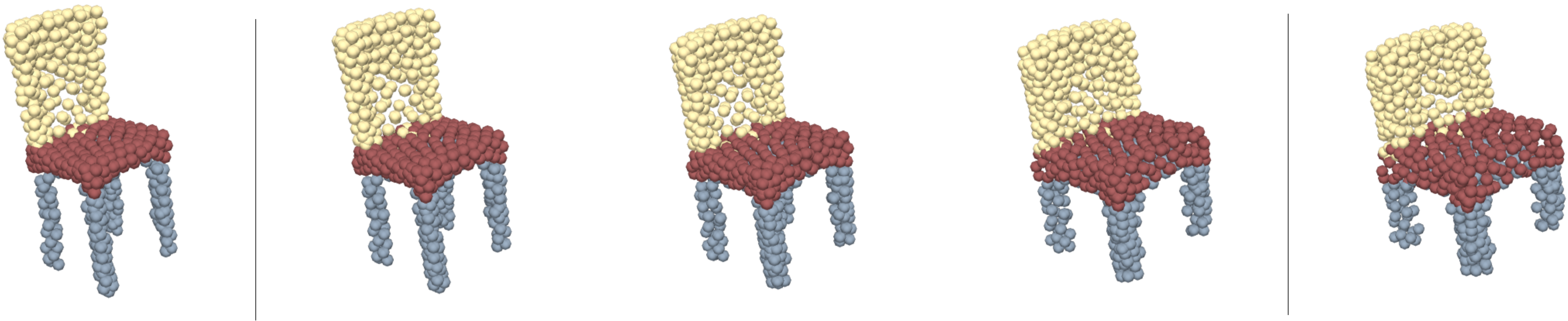}
    \end{center}
    \vspace{-3mm}
       \caption{Interpolation result. 
       Leftmost to rightmost by mixing latents with weights 0.2, 0.5, 0.8, respectively.
       }
    \label{fig:interpolation}
\end{figure}%
Two generated point clouds are interpolated by first mixing corresponding latents with different weights, and then pass it to corresponding generators.
The visualization results is shown in~\figref{fig:interpolation}.
As we can see, the middle three point clouds are deforming continuously from the leftmost to rightmost.
Thus, the learned latent space is continuous.

\section{Semantic meaningfulness}
Note that the {\bf Arxiv} paper \mrgan~\cite{gal2020mrgan} lacks accompanying code, we only compared semantic meaningfulness with \treegan~\cite{shu20193d} quantitatively in the main paper.
Here we show the qualitative comparison with \mrgan~\cite{gal2020mrgan} via \textit{their} main Figure 3 and supplmentary Figure 1:
For example, \mrgan's table bases are separated into three parts, some of them even linked to a leg, while \textsc{EditVAE} separates base and legs more clearly as per rows 2-3 in the main paper Figure 3.

\section{Superquadrics visualization}
\begin{figure}[t]
    \begin{center}
    \vspace{-3mm}
       \includegraphics[width=1\linewidth]{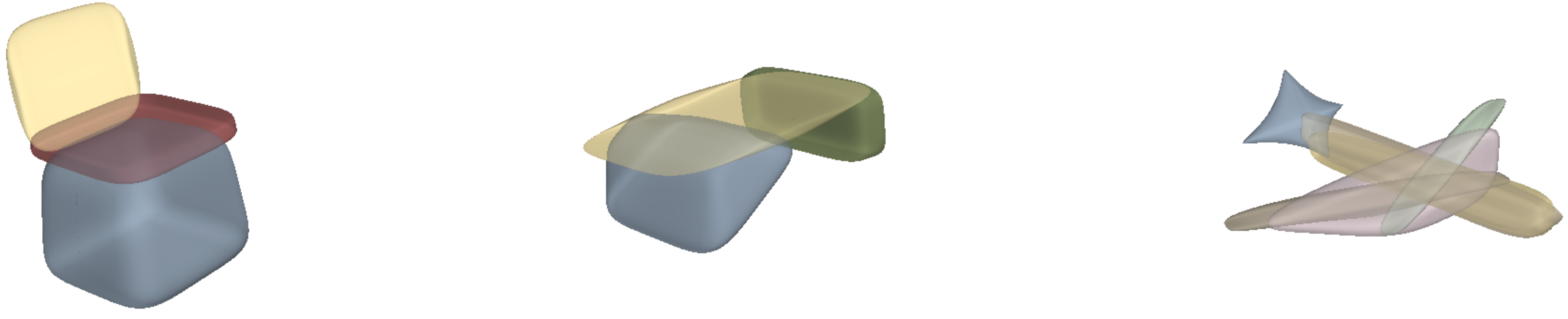}
    \end{center}
    \vspace{-3mm}
       \caption{Generated Superquadrics. Listed chair's $M=3$, table's $M=5$, airplane's $M=6$ as in Table 1 in main paper.  
       }
       \vspace{-6mm}
    \label{fig:superquad}
\end{figure}%
See \figref{fig:superquad} for an example for the generated superquadrics by passing sampled latents into pose and primitive branches in the main paper Figure 3.

\section{Primitive Detail}\label{sect:prim_detail}
\subsection{Deformation}
As mentioned in Preliminaries of the main paper, we use a tapering deformation~\cite{barr1987global} to enhance the representation power of the superquadrics.
Following the code provided by~\citet{paschalidou2019superquadrics}, the tapering deformation $\bm K$ is defined by:
\begin{align}
    \bm K(\bm x) = 
    \begin{bmatrix}
        \frac{k_1}{\alpha_z}x_z & 0 & 0\\
        0 & \frac{k_2}{\alpha_z}x_z & 0\\
        0 & 0 & 1 
    \end{bmatrix}
    \begin{bmatrix}
        x_x\\
        x_y\\
        x_z
    \end{bmatrix},
\end{align}
where $\bm x = (x_x, x_y, x_z)^\top$ is a point, and $\bm k = (k_1, k_2)^\top$ defines deformation parameters, and $\alpha_z$ is the size parameter in the $z$-axis.
This linear taper deforms the primitive shape in the $x,y$-axis by an amount which depends on the value of the $z$-axis.
As a result, tapering deformation will make primitives more conic, which helps to model unbalanced shapes such as the head of the airplane.

\subsection{Details on the Superquadric Losses}
While the definition our superquadric loss functions follows~\citet{paschalidou2019superquadrics}, we include more details here for the sake of completeness.

The superquadric loss is defined as
\begin{align}
    \mathcal L_s(\bm P, \bm X) = \mathcal L_D(\bm P, \bm X) + \mathcal L_r(\bm P),
\end{align}
where $\mathcal L_D$ is the distance term which encourages superquadric $\bm P$ to fit the input point cloud $\bm X$.
$\mathcal L_r$ is a regularisation term which encourages desired behaviour;
for example, we prefer primitives that do not overlap one another.

\noindent{\bf The distance term} measures the distance between points sampled from primitive surface and input point cloud $\bm X$.
Following the idea of the Chamfer distance, the distance term is decomposed by:
\begin{align}
    \mathcal{L}_D(\mathbf{P}, \mathbf{X}) = 
    \mathcal{L}_{\mathbf{P}\rightarrow \mathbf{X}}(\mathbf{P}, \mathbf{X}) 
    + \mathcal{L}_{\mathbf{X}\rightarrow \mathbf{P}}(\mathbf{X}, \mathbf{P}),
\end{align}
where $\mathcal{L}_{\mathbf{P}\rightarrow \mathbf{X}}$ defines the distance from the primitive $\mathbf{P}$ to the input point cloud $\mathbf{X}$,
 and $\mathcal{L}_{\mathbf{X}\rightarrow \mathbf{P}}$ defines the distance from the point cloud $\mathbf{X}$ to primitive $\mathbf{P}$.
Additional details may be found in~\cite{paschalidou2019superquadrics}.

\noindent{\bf The regularisation term} is defined as
\begin{align}
    \mathcal L_r(\bm P) = 
    \omega_o \mathcal L_o(\bm P).
\end{align}
As we manually select the number of parts, we only use an \textbf{overlapping regularizer} to discourage the superquadrics from overlapping one another; this term is adapted from~\cite{paschalidou2019superquadrics}.

In order to achieve the best performance, different $\omega_o$ are used for different categories during training. In particular 
we set: $\omega_o=1e-6$ for the chair category with number of primitives $M=3$;
$\omega_0=1e-5$ for the chair category with $M=7$, and the airplane category with $M=3$ and $M=6$ ;
$\omega_o=1e-10$ for the table category with $M=3$ and $M=5$. 

\section{Model details}
\label{sec:modeldetails}
We give two alternative derivations of our training objective, followed by some additional discussions, and details of our network architectures.

\subsection{Detailed Derivations}
To make the supplementary material self contained, we first recall inequality (7) in the main paper,
\begin{align}
    \hspace{-1.5mm} \log P_\theta(\bm X )
    & \geq \int 
    Q_\phi(\bm z, \bm \zeta|\bm X)\log \frac{P_\theta(\bm X, \bm z, \bm \zeta ) }{Q_\phi(\bm z, \bm \zeta|\bm X)}
    \intd \bm z \intd \bm \zeta,
    \label{eqn:hardelbo}
\end{align}
as well as equations (8) and (9) in the main paper,
\begin{align}
    & P_\theta(\bm z, \bm \zeta | \bm X) = P_\theta(\bm \zeta|\bm z) P_\theta(\bm z | \bm X),
    \label{eqn:pfactors}\\
    & Q_\phi(\bm z, \bm \zeta | \bm X)
    \equiv
    \label{eqn:qexactfactors}
    P_\theta(\bm \zeta | \bm z) \, Q_\phi(\bm z | \bm X).
\end{align}
\subsubsection{First derivation}
\label{subsect:variantion_inference}
By putting \eqref{eqn:pfactors} and \eqref{eqn:qexactfactors} into the lower bound \eqref{eqn:hardelbo}, we have
\begin{align}
    &\log P_\theta(\bm X) \geq \\
    &\int 
    P_\theta(\bm \zeta | \bm z) \, Q_\phi(\bm z | \bm X)\log 
    \frac{P_\theta(\bm \zeta|\bm z) P_\theta(\bm z | \bm X) P_\theta(\bm X) }{P_\theta(\bm \zeta | \bm z) \, Q_\phi(\bm z | \bm X)}
    \intd \bm z \intd \bm \zeta.
    \nonumber
\end{align}
By cancelling $P_\theta(\bm \zeta | \bm z)$ and taking the integral of $\bm \zeta$ we get
\begin{align}
    \log P_\theta(\bm X)
    \geq \int 
    Q_\phi(\bm z | \bm X)\log 
    \frac{ P_\theta(\bm z | \bm X) P_\theta(\bm X) }{Q_\phi(\bm z | \bm X)}
    \intd \bm z.
\end{align}
By applying Bayes' rule, we have
\begin{align}
    \log P_\theta(\bm X)
    & \geq \int 
    Q_\phi(\bm z | \bm X)\log 
    \frac{ P_\theta(\bm X | \bm z) P_\theta(\bm z) }{Q_\phi(\bm z | \bm X)}
    \intd \bm z \\
    & = \int Q_\phi(\bm z | \bm X)\log P_\theta(\bm X | \bm z)
    \intd \bm z \nonumber \\
    &\quad - \int \label{eqn:kl}
    Q_\phi(\bm z | \bm X)\log 
    \frac{Q_\phi(\bm z | \bm X) }{P_\theta(\bm z)}
    \intd \bm z.
\end{align}
We see the key point, that the final term in \eqref{eqn:kl} is tractable as it does not depend on $\bm \zeta$, that is $D_{KL}\left( Q_\phi(\bm z | \bm X)) \Vert P_\theta(\bm z) \right)$.
Since our decoder has a simple deterministic relationship which we denote by the limit 
\begin{align}
    Q_\phi(\bm \zeta|\bm z)\rightarrow\delta(\bm \zeta - \mathrm{NN}_\theta(\bm z)),
\end{align}
we can rewrite the reconstruction error term to emphasise the dependence of $\bm X$ on $\bm \zeta$
to get the ELBO
\begin{align}
    \log P_\theta(\bm X)  
    & \,\,\,\geq\,\,\, \mathbb E_{Q_\phi(\bm z |\bm X)}
    \left[
    \log P_\theta(\bm X|\bm \zeta)
    \right] \nonumber\\
     &\quad-D_{KL}\left( Q_\phi(\bm z | \bm X)) \Vert P_\theta(\bm z) \right),
\end{align}
where $\bm \zeta = \mathrm{NN}_\theta(\bm z)$.

\subsubsection{Second Derivation}
\label{subsect:variantion_inference}
Using \eqref{eqn:pfactors} and \eqref{eqn:qexactfactors} in \eqref{eqn:hardelbo}, we have:
\begin{align}
    \log P_\theta(\bm X)
    & \geq 
     \int 
    Q_\phi(\bm z, \bm \zeta| \bm X)\log 
    \frac{P_\theta(\bm X, \bm z, \bm \zeta)}{P_\theta(\bm \zeta | \bm z) \, Q_\phi(\bm z | \bm X)}
    \intd \bm z \intd \bm \zeta
    \nonumber\\
    & ~~~~ = \int 
    Q_\phi(\bm z, \bm \zeta| \bm X)\log 
    \frac{P_\theta(\bm X|\bm z, \bm \zeta)P_\theta(\bm z, \bm \zeta)}{Q_\phi(\bm z, \bm \zeta | \bm X)}
    \intd \bm z \intd \bm \zeta
    \nonumber\\
    & ~~~~ = 
    \underbrace{
    \mathbb E_{Q_\phi(\bm z, \bm \zeta |\bm X)}
    \left[
    \log P_\theta(\bm X|\bm z, \bm \zeta)
    \right]}_L
    \nonumber \\
    &\quad \quad-
    \underbrace{
    \mathbb E_{Q_\phi(\bm z, \bm \zeta |\bm X)}
    \left[
    \log \frac{Q_\phi(\bm z, \bm \zeta|X)}{P_\theta(\bm z, \bm \zeta)}
    \right]
    }_R.
\end{align}
The key point is revealed, that the regulariser term is tractable because, by \eqref{eqn:qexactfactors}
\begin{align}
R
& =
    \mathbb E_{Q_\phi(\bm z, \bm \zeta |\bm X)}
    \left[
    \log \frac{Q_\phi(\bm z, \bm \zeta|X)}{P_\theta(\bm z, \bm \zeta)}
    \right]    
\\
& =
    \mathbb E_{Q_\phi(\bm z, \bm \zeta |\bm X)}
    \left[
    \log \frac{P_\theta(\bm \zeta | \bm z)Q_\phi(\bm z|\bm X)}{P_\theta(\bm \zeta|\bm z)P_\theta(\bm z)}
    \right]
\\
& =
    \mathbb E_{Q_\phi(\bm z, \bm \zeta |\bm X)}
    \left[
    \log \frac{Q_\phi(\bm z|\bm X)}{P_\theta(\bm z)}
    \right]  
\\
& =
    \mathbb E_{Q_\phi(\bm z|\bm X)}
    \left[
    \log \frac{Q_\phi(\bm z|\bm X)}{P_\theta(\bm z)}
    \right]
    \\
& = 
D_{KL}\left( Q_\phi(\bm z | \bm X) \Vert P_\theta(\bm z) \right).
\end{align}
Finally, since our decoder has a simple deterministic relationship which we denote by the limit 
\begin{align}
    Q_\phi(\bm \zeta|\bm z)\rightarrow\delta(\bm \zeta - \mathrm{NN}_\theta(\bm z)),
\end{align}
we can rewrite the reconstruction error term to emphasise the dependence of $\bm X$ on $\bm \zeta$,
\begin{align}
L & =
\mathbb E_{Q_\phi(\bm z, \bm \zeta |\bm X)}
    \left[
    \log P_\theta(\bm X|\bm z, \bm \zeta)
    \right]    
\\
& =
\mathbb E_{Q_\phi(\bm z|\bm X)}
    \left[
    \log P_\theta(\bm X|\bm \zeta)
    \right],
\end{align}
where in the final line $\bm \zeta = \mathrm{NN}_\theta(\bm z)$. 
\subsection{Discussion}
Because of the deterministic mapping between $\bm \zeta$ and $\bm z$, we have $P_\theta(\bm \zeta | \bm z) = Q_\theta(\bm \zeta | \bm z)$. This allows us to annotate $P_\theta(\bm \zeta | \bm z)$ as in Figure~2 of the main paper.

In that same figure, we annotate with $P_\theta(\bm X | \bm \zeta)$ the mapping (on the right hand side) from parts representations $\bm \zeta$ to the output point cloud $\bm Y$, despite $\bm X$ appearing on the left hand side. This is consistent with standard expositions: for example, we may connect this with Appendix~C.1 of the VAE paper~\cite{kingma2013auto} by noting that our $\bm Y$ and $\bm P$ are together analogous to their Bernoulli parameters $\bm p$.

Finally note that the posterior of, for example, the combined primitive $\bm P$ is not included in our variational inference model, which is a byproduct obtained  by assembling the part primitives from posterior samples of $\bm \zeta$.

\subsection{Architecture and Implementation Details}\label{subsect:network_detail}
The model is implemented using PyTorch~\cite{paszke2019pytorch} on the platform of Ubuntu 16.04, trained on one GeForce RTX 3090 and one GeForce RTX 2080 TI.
10 Gigabytes memory is allocated.

The number of parts (parameter $M$) for each category is manually selected with domain specific knowledge.
The choice reflects what one believes a good semantic segmentation should be, which is application dependent.
As mentioned in the main paper, a chair may be roughly decomposed into back, base and legs for $M=3$.
In addition, a chair could also be decomposed into back, base, armrest, and four legs for $M=7$.
For the airplane category, it could separated into body, tail, and wings for $M=3$; and also into body, two wings, two engines, and tail for $M=6$.
Finally, a table may be decomposed into base and four legs for $M=5$; and also into base, left two legs, and right two legs for $M=3$.
The general message here is clear: various decompositions are valid and useful.

\paragraph{Encoder}
We use \textsc{PointNet}~\cite{qi2017pointnet} as the encoder.
Following the network structure from~\cite{achlioptas2018learning}, the encoder has 64, 128, 128, 256 filters at each layer.
Batch normalization and \textsc{LeakyReLU} are used between each layer.

\paragraph{Point Decoder}
We use the generator of TreeGAN~\cite{shu20193d} as the point decoder.
The architecture we used has 5 layers, with the root being the local part latent vector, and the leaves being points in $\mathbb R^3$.
The loop term has $K=10$ supports.
The feature dimension and branching factor for each layer are $[32, 32, 16, 16, 3]$ and $[1, 2, 4, 32]$, respectively.
Hence, each point decoder outputs $256$ points.

\paragraph{Pose and Primitive Decoders}
All pose and primitive decoders are one layer fully connected networks, following~\cite{paschalidou2019superquadrics}.
The dimension of the fully connected layers depends on the input latent size (namely 8) and output parameter dimension.
See the repository of~\cite{paschalidou2019superquadrics} for the detailed implementation.

\end{document}